\documentclass[preprint,12pt]{elsarticle}

\usepackage{amssymb}
\usepackage[utf8]{inputenc}
\usepackage{soul}
\usepackage{color}
\usepackage{graphicx}
\usepackage{amsmath}
\usepackage[version=4]{mhchem}
\usepackage{siunitx}
\usepackage{longtable,tabularx}
\usepackage{caption}
\usepackage{float}
\usepackage[nameinlink,capitalise,noabbrev]{cleveref}
\usepackage[dvipsnames]{xcolor}
\usepackage{algorithm}
\usepackage{algorithmic}
\graphicspath{{Figures/}}
\usepackage{subfigure}
\usepackage{multirow}
\usepackage{booktabs}
\usepackage{xcolor}
\usepackage{geometry}
\usepackage{array}
\usepackage{textcomp}
\usepackage{ulem}
\usepackage{placeins}

\journal{arXiv}

\begin{document}

\begin{frontmatter}

\title{Generative Spatio-temporal GraphNet for Transonic Wing Pressure Distribution Forecasting}

\author[inst1,inst2]{Gabriele Immordino\corref{cor1}}
\ead{G.Immordino@soton.ac.uk}

\fntext
[label1]{Ph.D. Student}
\cortext[cor1]{Corresponding Author}

\affiliation[inst1]{organization={Faculty of Engineering and Physical Sciences, University of Southampton},
            % addressline={176/5049 Boldrewood Innovation Campus}, 
            city={Southampton},
            % postcode={SO17~1BJ}, 
            country={United Kingdom}}

\author[inst2]{Andrea Vaiuso\fnref{label2}}
\fntext[label2]{Research Associate}

\author[inst1]{Andrea Da Ronch\fnref{label3}}
\fntext[label3]{Professor, AIAA Senior Member}

\author[inst2]{ Marcello Righi\fnref{label4}}
\fntext[label4]{Professor, AIAA Member, Lecturer at Federal Institute of Technology Zurich ETHZ}

\affiliation[inst2
]{organization={School of Engineering, Zurich University of Applied Sciences ZHAW},
            % addressline={Technikumstrasse 9}, 
            city={Winterthur},
            % postcode={8400}, 
            country={Switzerland}}

\begin{abstract}

This study presents a framework for predicting unsteady transonic wing pressure distributions, integrating an autoencoder architecture with graph convolutional networks and graph-based temporal layers to model time dependencies. The framework compresses high-dimensional pressure distribution data into a lower-dimensional latent space using an autoencoder, ensuring efficient data representation while preserving essential features. Within this latent space, graph-based temporal layers are employed to predict future wing pressures based on past data, effectively capturing temporal dependencies and improving predictive accuracy. This combined approach leverages the strengths of autoencoders for dimensionality reduction, graph convolutional networks for handling unstructured grid data, and temporal layers for modeling time-based sequences. The effectiveness of the proposed framework is validated through its application to the Benchmark Super Critical Wing test case, achieving accuracy comparable to computational fluid dynamics, while significantly reducing prediction time. This framework offers a scalable, computationally efficient solution for the aerodynamic analysis of unsteady phenomena.

% The feedforward model, in particular, outperforms the autoregressive approach in long-term predictions, effectively mitigating error propagation.

\end{abstract}

\end{frontmatter}

\section*{Nomenclature}

\noindent
\begin{minipage}[t]{0.4\textwidth} % Left column for Acronyms
    \textbf{Acronyms}
    {\renewcommand\arraystretch{1.0}
    \begin{longtable}{@{}p{1.4cm} @{\quad=\quad} p{7cm}@{}}
    $AE$ & autoencoder \\
    $CFD$ & computational fluid dynamics \\
    $GCN$ & graph convolutional network \\
    $GNN$ & graph neural network \\
    $GRU$ & gated recurrent unit \\
    $GST$ & generative spatio-temporal \\
    $LSTM$ & long short-term memory \\
    $ML$ & machine learning \\
    $MAE$ & mean absolute error \\
    $MAPE$ & mean absolute percentage error \\
    $MWLS$ & moving weighted least squares \\
    $RMSE$ & root mean square error \\
    $ROM$ & reduced-order model \\
    $STGCN$ & spatio-temporal GCN \\
    \end{longtable}}
\end{minipage}%
\hfill % Space between columns
\begin{minipage}[t]{0.4\textwidth} % Right column for Symbols
    \textbf{Symbols}
    {\renewcommand\arraystretch{1.0}
    \begin{longtable}{@{}p{0.7cm} @{\quad=\quad} p{7cm}@{}}
    % $AoA$   & angle of attack, deg \\
    % $c$   & mean chord, m \\
    % $C_{D}$ & drag coefficient \\
    % $C_{F}$ & skin friction coefficient \\
    $C_{L}$ & lift coefficient \\
    $C_{M}$ & pitching moment coefficient \\
    $C_{P}$ & pressure coefficient \\
    $M$ & Mach number \\
    $\theta$ & pitch angle, rad \\
    $\dot{\theta}$ & pitch rate, rad/s \\
    $\ddot{\theta}$ & pitch acceleration, rad/s$^2$ \\
    $\xi$ & plunge, m \\
    $\dot{\xi}$ & plunge rate, m/s \\
    $\ddot{\xi}$ & plunge acceleration, m/s$^2$ \\
    % $t$ & time, s \\
    \end{longtable}}
\end{minipage}

\section{Introduction}

The complexity of aerodynamic analysis poses a significant challenge across various engineering applications. Accurately predicting challenging physical phenomena involves capturing detailed variations that arise from the complex interaction of multiple forces. Traditional computational fluid dynamics (CFD) methods are effective in many scenarios, but often require substantial computational resources and may struggle with accurately representing dynamic and unsteady phenomena under specific flow conditions~\cite{anderson1995computational, blazek2015computational}. These limitations highlight the need for more efficient and robust approaches.

Recent advancements in machine learning (ML) offer promising solutions for these challenges. Initially, deep neural networks were applied to capture complex patterns in fluid dynamics~\cite{sabater2022fast, castellanosassessment, immordino2023steady, tompson2017accelerating}. However, aerospace engineering problems often rely on non-homogeneous and unstructured grids modeling, which necessitate more advanced ML architectures that can handle this complex data structures. 

% Indeed, the CFD method entails the discretization of the fluid domain into a computational grid, commonly referred to as a mesh. This is often achieved through the use of three-dimensional unstructured geometries, which enables the attainment of enhanced resolution in complex regions. 

Geometric deep learning, introduced around 2017~\cite{bronstein2017geometric}, utilizes graph neural networks (GNNs) for graph-structured data~\cite{gori2005new, scarselli2009graph}. GNNs excel in capturing relationships and dependencies within graph nodes, making them ideal for tasks involving topological information~\cite{wu2020comprehensive, zhang2020deep, zhou2020graph}. Graph Convolutional Networks (GCNs), a specific type of GNN, leverage convolution operations on graphs to enhance their capability to process graph-structured data~\cite{kipf2016semi}. GCNs are particularly promising in aerospace engineering, as they can handle data with spatial structures and are suitable for modeling complex aerodynamic geometries~\cite{jin2018prediction, fukami2019super, omata2019novel, peng2020time, rozov2021data}. In fact, while Convolutional Neural Networks (CNNs) perform well on regular grid data like images and texts, GCNs are better suited for irregular domains, such as mesh grids, by applying convolution operations directly on graphs~\cite{kipf2016semi}. Indeed, GCNs can directly input raw 3D model mesh data without pre-computation or feature extraction, enhancing predictive capabilities without bias or information loss~\cite{wu2020comprehensive}.

Another important challenge concerns the high dimensionality of model input data. As with CNN architectures for image recognition tasks, deep and complex architectures struggle with propagating information over a large number of features. CFD simulations typically involve the use of fine meshes, consisting of a significantly large number of points, increasing both complexity and computational requirements. To manage this, careful data compression is needed to retain only the essential features without losing critical information. Previous studies have demonstrated that dimensionality reduction can be effectively achieved using an autoencoder (AE) architecture ~\cite{han2019novel,rozov2021data,saetta2022machine,massegur2023graph}. AEs, through their encoding and decoding processes, can learn a compact and efficient representation of the data, ensuring that critical information is preserved while reducing the computational burden~\cite{hinton2006reducing, vincent2008extracting}. 

Building on our previous study~\cite{immordino2024predicting}, which focused on steady-state problems, we now extend our methodology to address unsteady phenomena. Predicting time-varying pressure distributions relies primarily on capturing temporal dependencies within the data. Recurrent Neural Networks (RNNs), with their ability to track evolving patterns through a hidden state, are particularly well-suited for this task. Their effectiveness in modeling unsteady behaviors and dynamic responses makes them an ideal choice for forecasting time series in aerodynamic applications~\cite{hochreiter1997long, cho2014learning}. However, RNNs often struggle with long-term dependencies due to challenges like vanishing gradients~\cite{salehinejad2017recent}. To address these limitations, Long Short-Term Memory (LSTM) networks and Gated Recurrent Units (GRUs), both RNN variants, have been developed. LSTMs introduce gates that control the flow of information, making them more effective at learning long-term dependencies~\cite{hochreiter1997long}. GRUs offer a simpler structure than LSTMs, using fewer gates while still managing to capture long-term dependencies, often with faster training times~\cite{cho2014learning}. LSTMs have been extensively applied in aerodynamic modeling, such as predicting the dynamic response of aeroelastic systems and turbulence~\cite{li2019deep, brunton2020machine}, while GRUs have also proven effective for similar tasks~\cite{wang2020multivariate, mannarino2014nonlinear}. More recently, attention mechanisms have revolutionized time series forecasting by enabling models to focus on the most relevant parts of the input sequence~\cite{vaswani2017attention, cheng2016long}. This capability leads to more accurate and robust predictions, as demonstrated for instance by improvements in maintenance scheduling through estimating icing probabilities on wind turbine blades~\cite{cheng2021temporal}, stable long-term fluid dynamics predictions using transformer-style temporal attention~\cite{han2022predicting}, and enhanced design and control of hypersonic vehicles by capturing spatiotemporal turbulence characteristics~\cite{du2024novel}. Attention mechanisms enable models to weigh the importance of different time steps dynamically, thereby improving the ability to model complex temporal patterns. Similarly, Spatio-Temporal Graph Convolution Networks (STGCNs) have shown strong performance in modeling such patterns by processing entire sequences in parallel and applying filters across the time dimension, capturing both short- and long-term dependencies efficiently~\cite{bai2021a3t}.

To fully harness the potential of these temporal modeling techniques, it is important to recognize that each approach offers distinct benefits depending on the nature of the data and the specific application. With this in mind, our methodology investigates and evaluates several temporal layers— LSTMs, GRUs, attention mechanisms, and STGCNs—to effectively capture the temporal dependencies in our case study. By integrating graph convolutional networks with AEs and temporal layers, our proposed approach leverages the strengths of each method to enhance the prediction of unsteady surface pressure distributions over a transonic wing. This integrated framework not only handles the unstructured grids typical in aerodynamic data through GCNs but also ensures efficient dimensionality reduction through AEs. By comparing different temporal approaches, we aim to provide a comprehensive and scalable solution for complex aerodynamic analyses, delivering more accurate and computationally efficient predictions for unsteady phenomena.

The structure of the paper is as follows: Section~\ref{sec:methodology} details the implemented methodology, providing a comprehensive explanation of the architecture and its components. Section~\ref{sec:test_case} describes the test case used to validate the model, focusing on its aerodynamic challenges. Section~\ref{sec:results} presents the results, comparing the performance of different architectures designed to address temporal challenges, while evaluating the impact of various temporal layers. Finally, Section~\ref{sec:Conclusions} summarizes the conclusions drawn from the study and suggests potential future directions for improving the model accuracy and scalability in different aerodynamic scenarios.

\section{Methodology}\label{sec:methodology}

This section outlines the methodology used in developing the model. It begins with an overview of the Generative Spatio-Temporal (GST) GraphNet framework, then provides an in-depth description of each component of the model.

\subsection{Generative Spatio-Temporal GraphNet}

The proposed GST GraphNet framework integrates a GCN-based AE architecture with a temporal prediction layer to effectively model and forecast wing pressure distributions for subsequent timesteps. The encoding and decoding modules operate with graph nodes based on the pressure-gradient distribution values across the wing surface, performing pooling and unpooling operations on the input, respectively. Initially, a pre-trained AE is used to compress the pressure distribution data into a lower-dimensional representation, preserving fundamental features while reducing computational complexity. This pre-training step reduces the full model training time and computational costs, enhancing the overall efficiency of the prediction process. After the pooling operation, the reduced-space representation is passed through a temporal prediction layer. This layer is designed to capture the temporal dependencies in the data and forecast wing pressure from a series of previous timesteps to a future one. To account for the complexities caused by shock waves and boundary layer separation, which affect force and moment calculations, a penalty term for the pitching moment coefficient $C_{M_y}$ is added to the Mean Absolute Error (\texttt{MAE}) loss function. This addition is represented as $Loss = \texttt{MAE} + \lambda \cdot C_{M_y}$, with $\lambda = 0.01$ for dimensional consistency. The combination of AE, GCN layers, and temporal modeling enables the framework to provide precise and reliable pressure predictions, which are crucial for analyzing aerodynamic performance.

\begin{figure}[!htb] 
    \centering
    \includegraphics[width=\textwidth]{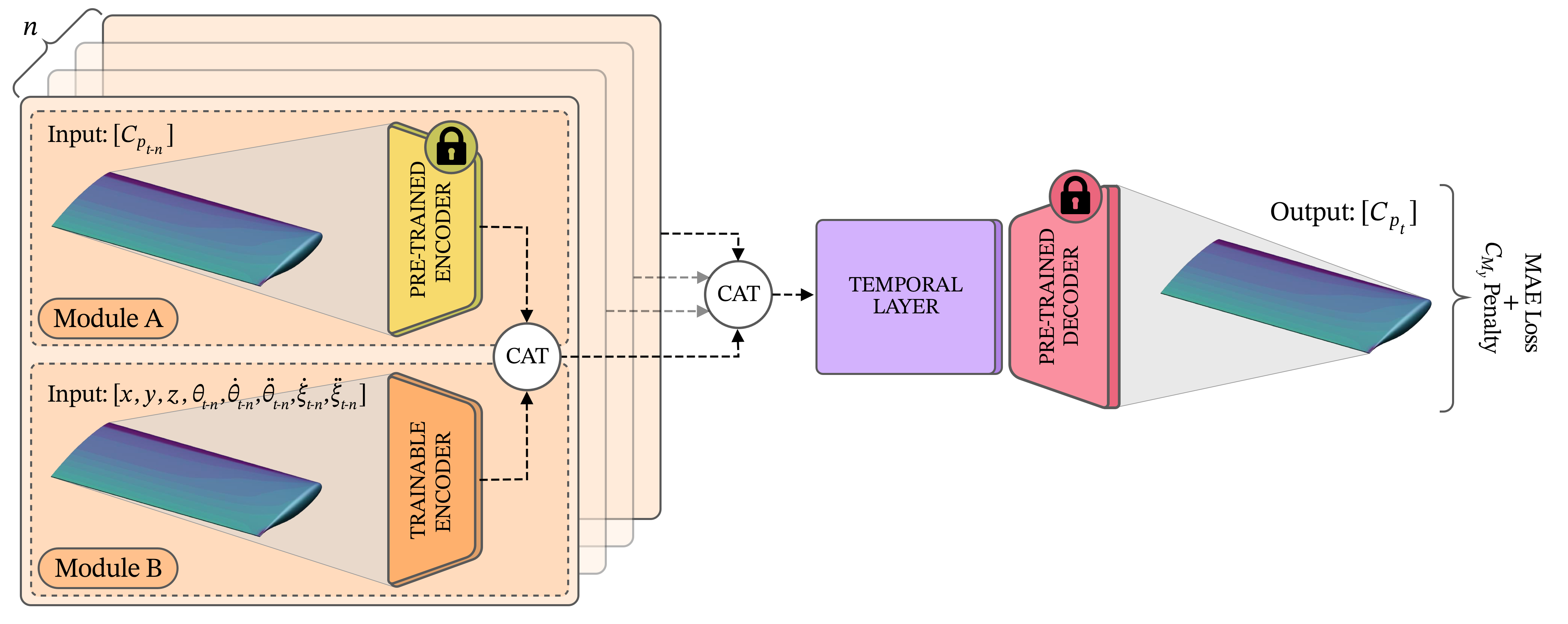}
    \caption{Overview of the GST GraphNet architecture for predicting wing pressure distributions. Module A represents the autoregressive component, incorporating previously predicted \(C_P\) values, while Module B processes spatial coordinates and motion data from previous timesteps.}
    \label{fig:architecture}
\end{figure}

A visual overview of the model architecture is presented in Figure~\ref{fig:architecture}. The model input features include data from the $n$ previous timesteps (\(t-1\), ...,  \(t-n\)): spatial coordinates (\(x, y, z\)); pitch (\(\theta_{t-n}, \dot{\theta}_{t-n}, \ddot{\theta}_{t-n}\)); plunge (\(\dot{\xi}_{t-n}, \ddot{\xi}_{t-n}\)); and pressure coefficient (\(C_{P_{t-n}}\)), with \(n =3\). Finally, the output of the model is represented by the pressure coefficient (\(C_{P_{t}}\)) at the current timestep \(t\). The choice of this sequence length ensures that the model has access to sufficient temporal context to capture the evolution of unsteady aerodynamic features, such as flow separation and shock dynamics, while avoiding the inclusion of redundant or excessive data, which would increase computational complexity without significantly improving accuracy.

Building on this, we developed two different types of architecture: a feedforward model and an autoregressive–moving-average model with exogenous inputs (ARMAX) model. In the case of the feedforward model, the inputs consist solely of the coordinates of the wing surface and the prescribed motion at \( n \) previous timesteps casted on each graph node (using only Module B in Figure~\ref{fig:architecture}). This model does not incorporate any past predicted pressures into its input, relying purely on the historical spatial and motion data of the wing to make its predictions. Conversely, the ARMAX model includes additional information in its input by integrating the pressures predicted at prior timesteps (utilizing both Module A and Module B in Figure~\ref{fig:architecture}). This autoregressive component allows the ARMAX model to potentially capture more complex temporal dependencies by considering the history of its own predictions, aiming to enhance the accuracy of the pressure forecasts. Implementing both models serves to evaluate the trade-offs between simplicity and predictive depth: while the feedforward model offers a simpler, stable approach less prone to error accumulation, the ARMAX model is designed to capture complex temporal dependencies and unsteady behaviors, potentially enhancing accuracy under dynamic conditions.

To limit error accumulation in the time-marching scheme, we employed a Back-Propagation Through Time (BPTT) algorithm~\cite{salehinejad2017recent} for the total loss calculation, dividing the dataset into mini-sequences. The model processes each sequence consecutively, accumulating error over time. After processing each sequence, the loss function is applied to update the model parameters through backpropagation. A sequence length of three was chosen based on its performance, yielding the best results.

\subsubsection{Graph Representation}

A graph \( G \) is defined by a set of nodes \( N \) and edges \( E \), where each edge \( (i, j) \) represents a directed link from node \( i \) to node \( j \). Self-loops occur when \( (i,i) \in E \). These connections are represented by an adjacency matrix \( \mathbf{A} \), where \( \mathbf{A}_{ij} = 1 \) if \( (i,j) \in E \), and 0 otherwise. Costs for edges can be included by replacing \( 1 \) with the cost and using \( \infty \) for absent connections. A path \( p(i \rightarrow j) \) is a series of steps from \( i \) to \( j \), where each step \( (h,k) \in E\). A graph is acyclic if no path returns to a starting node, otherwise, it is cyclic.

In our case, the mesh can be represented as a cyclic graph \( G \), where each grid point \( i \) serves as a node. Each node carries features such as spatial coordinates \((x, y, z)\), pitch \((\theta, \dot{\theta}, \ddot{\theta})\), plunge \((\dot{\xi}, \ddot{\xi})\), and the pressure coefficient \( C_P \) from the previous \( n \) timesteps. We call node features \(y_i\), while edge weights \(e_{ij}\).

Graph connectivity is represented by the adjacency matrix \(\mathbf{A}\), where each weight \(e_{ij}\) is the Euclidean distance between grid points \(i\) and \(j\): \(e_{ij} = \|\mathbf{x}_i - \mathbf{x}_j\|_2\). To normalize the weights to \((0, 1]\) and include self-loops (\(e_{ii} = 1\)), the adjacency matrix is augmented: \( \mathbf{\Tilde{A}} = \mathbf{A} + \mathbf{I}\). Since edges are bidirectional (\(e_{ij} = e_{ji}\)), \( \mathbf{\Tilde{A}} \) is symmetric.

Due to the graph sparsity, the adjacency matrix is stored in a memory-efficient Coordinate List (COO) format, where the edge-index matrix contains pairs of node indices, and the edge-weight matrix stores the corresponding weights, with \(n_e\) being the number of edges.

\subsubsection{Graph Convolutional Networks}

GCNs are a particular type of ML algorithms that are based on graph theory. GCNs extract features from graphs by aggregating information from neighboring nodes using a graph convolutional operator. This operator, originally introduced by Duvenaud et al. in 2015 for molecular fingerprints~\cite{duvenaud2015convolutional}, was later extended by Kipf et al. in 2016~\cite{kipf2016semi} and is now implemented in \texttt{PyTorch-Geometric} library~\cite{Fey/Lenssen/2019}. GCNs effectively generate node embeddings that capture structural information, making them ideal for tasks requiring an understanding of relationships between nodes.

The GCN operator follows the layer-wise propagation rule that is defined by the equation:

\begin{equation}\label{eq:gcn}
    H^{(l+1)} = \sigma (\mathbf{\Tilde{D}}^{-\frac{1}{2}}\mathbf{\Tilde{A}}\mathbf{\Tilde{D}}^{-\frac{1}{2}}H^{(l)}W^{(l)})
\end{equation}

Here, \(H^{(l)}\) and \(H^{(l+1)}\) are the node feature matrices at layers \(l\) and \(l+1\), \( \mathbf{\Tilde{A}} \) is the adjacency matrix with self-loops, \( \mathbf{\Tilde{D}} \) is the degree matrix calculated on \( \mathbf{\Tilde{A}} \), \(W^{(l)}\) is the matrix of trainable weights, and \( \sigma \) is the activation function. This rule propagates information from a node to its neighbors, allowing nodes to gather information from larger neighborhoods as layers are stacked.
Equation~\ref{eq:gcn} is a first-order approximation of trainable localized spectral filters \( g_{\theta} \) on graphs~\cite{kipf2016semi}.

A spectral convolution \( g_{\theta} \ast x \) of an input graph \( \Tilde{x} \) with a filter \( g_{\theta} \) in the Fourier domain is defined as:

\begin{equation}\label{eq:graph_conv_op}
    g_{\theta} \ast x = \mathbf{U} g_{\theta} \mathbf{U}^T x
\end{equation}

where \( \mathbf{U} \) contains the eigenvectors of the graph Laplacian, \( \mathbf{L} \). By approximating the spectral filter \( g_{\theta} \) using Chebyshev polynomials~\cite{hammond2011wavelets}, GCNs perform efficient localized filtering on graph data. This approximation simplifies the convolution process, making it feasible for large-scale graphs while preserving the ability to extract meaningful node features.

\subsection{Temporal Layers}

In our framework, we explored various layers for temporal modeling: GRUs, LSTMs, attention mechanisms, and STGCN layers. Each method offers a distinct way to capture temporal dependencies, with varying level of complexity and performance suited to different contexts. In this section, we provide a brief overview of these methods, highlighting their key features and how they are integrated into our model. This comparison helps evaluate their effectiveness in handling temporal sequences.

\subsubsection*{Gated Recurrent Unit}
The combination of GCNs with GRU~\cite{cho2014learning} offers several key advantages when dealing with spatio-temporal data. GRUs are widely used and well-suited for modeling temporal dependencies, but they can not directly used with non-Cartesian domains like graphs, where spatial relationships are irregular. By incorporating graph convolution operators on GRUs, it is possible to improve the generalization capability of the model by replacing traditional convolution with a graph convolution, which can handle arbitrary graph structures and effectively learn from unstructured data.

Based on the approach in~\cite{NIPS2015_07563a3f} where recurrent networks for fixed grid-structured sequence are introduced, Seo et al.~\cite{seo2018structured} proposed a Graph Convolutional Recurrent Network (GCRN) architecture for building a generalized GRU that works with unstructured sequence. To generalize the model to graphs, the 2D convolution is replaced by the graph convolution operator, here $*_g$, introduced in (\ref{eq:graph_conv_op}).
In particular, GRU cell in GCRN is defined by:

\begin{equation*}
    \begin{aligned}
        z_t = \sigma(W_{xz} *_g \, x_t + U_{hz} *_g \, h_{t-1} + b_z) \\
        r_t = \sigma(W_{xr} *_g \, x_t + U_{hr} *_g \, h_{t-1} + b_r) \\
        \hat{h}_t = \phi(W_{xh} *_g \, x_t + U_{hh} *_g \, (r_t \odot h_{t-1}) + b_h) \\
        h_t = (1 - z_t) \odot h_{t-1} + z_t \odot \hat{h}_t
    \end{aligned}
\end{equation*}

Here, $W_{xz} *_g \, x_t$ refers to the graph convolution operation of $x_t$ with spectral filters which are functions of the graph Laplacian $L$ parametrized by $K$ Chebyshev coefficients. $z_t$ is the update gate vector, $r_t$ is the reset gate vector, $\hat{h}_t$ is the candidate activation vector, $h_t$ is the hidden state at time step $t$, $W$ and $U$ are the trainable weight matrices for the input and hidden states, respectively, $\sigma$ is the logistic sigmoid function, and $\phi$ is the hyperbolic tangent function (or other possible activation functions). The operator $\odot$ denotes the Hadamard product, while $b$ represents the biases.

\subsubsection*{Long Short-Term Memory}

LSTM networks~\cite{hochreiter1997long} are particularly useful when the data involves long-term dependencies, as they include memory units that can store information across multiple timesteps. This makes them potentially suitable to model unsteady aerodynamic flows where past behavior influences future forecasts over long periods of time. While both LSTM and GRU address the vanishing gradient problem and are designed to capture temporal relationships, LSTM includes additional memory structures that enable it to retain information over longer time periods, by sacrificing computational power and increasing the number of parameters. 
The implementation of a convolutional graph based LSTM follows a similar approach presented before with GRU~\cite{seo2018structured}, by creating a model that replaces the 2D convolution with the graph convolution operator $*_g$. In particular:

\begin{equation*}
    \begin{aligned}
        i_t = \sigma(W_{xi} *_g \, x_t + W_{hi} *_g \, h_{t-1} + w_{ci} \odot c_{t-1} + b_i) \\
        f_t = \sigma(W_{xf} *_g \, x_t + W_{hf} *_g \, h_{t-1} + w_{cf} \odot c_{t-1} + b_f) \\
        c_t = f_t \odot c_{t-1} + i_t \odot \phi(W_{xc} *_g \, x_t + W_{hc} *_g \, h_{t-1} + b_c) \\
        o_t = \sigma(W_{xo} *_g \, x_t + W_{ho} *_g \, h_{t-1} + w_{co} \odot c_t + b_o) \\
        h_t = o_t \odot \phi(c_t)
    \end{aligned}
\end{equation*}

Where $i_t$ is the input gate, $f_t$ the forget gate, $c_t$ the cell state, $o_t$ the output gate and $h_t$ the hidden state, which is the output of the LSTM at time step $t$. As before, $\sigma$ is the logistic sigmoid function, $\phi$ is the hyperbolic tangent function, $W$ represents the trainable weight matrix, and the support $K$ of the graph convolutional kernels is defined by the Chebyshev coefficients. This extension of the standard LSTM architecture enables the model to learn temporal dependencies while also taking into account the spatial structure of the data.

\subsubsection*{Attention Mechanisms}

% In our context, attention should allow the model to prioritize specific moments in the flow history that are most relevant to predicting future pressure distributions, thus improving theoretically accuracy in complex unsteady aerodynamic scenarios.

Incorporating attention mechanisms allows for dynamically assigning importance to different time steps in a temporal sequence, which is especially useful in graph-based models where both complex spatial and temporal dependencies must be captured. Following the work of Bai et al.~\cite{bai2021a3t}, attention mechanisms can be employed to re-weight the influence of hidden states of a GCRN across time, enabling the model to focus on the most relevant time points for prediction, rather than treating each equally. The model was constructed by combining GCN and GRU to compute both the spatial and temporal domains of the graph, by using the graph convolution operator $*_g$ introduced before. In addition, the attention mechanism is used to compute a context vector that selectively weighs the hidden states of the GCRN.

First, for each time step, the hidden states of the GRU \( h_t \) are passed through an attention layer, where attention scores \( \alpha_t \) are computed using a softmax function. These scores are then used to weigh the hidden states, resulting in the context vector \( C \), which captures the global variation information. 

In particular, given a series of hidden states calculated by the recurrent network for $T$ time steps: $H=\{h_1,h_2,...,h_T\}$, the attention weights $\alpha_t, 1<t<T$ are computed using a softmax function on the scores derived from a multilayer perceptron (MLP):

\[
e_t = w^{(2)}(w^{(1)} H + b^{(1)}) + b^{(2)}, \,\,\,
\alpha_t = \frac{\exp(e_t)}{\sum_{i=1}^{T} \exp(e_i)}
\]

where \(w^{(1)}\), \(w^{(2)}\), \(b^{(1)}\), and \(b^{(2)}\) are trainable weights and biases in the MLP. 
The context vector \( C \) is then calculated by the weighted sum and used for implementing the attention mechanism on the GCRN hidden states:

\[
\quad C = \sum_{t=1}^{T} \alpha_t h_t
\]

By combining GCNs for spatial feature extraction with GRUs and attention mechanisms for temporal modeling, the model can capture both short-term and long-term dependencies in the data.

\subsubsection*{Spatio-temporal Graph Convolution}

STGCN is introduced by Yu et al.~\cite{yu2017spatio} as a method to capture temporal dependencies in spatio-temporal data by applying convolutions across the time dimension. Unlike RNNs, which process inputs sequentially, temporal convolutions handle entire sequences at once, allowing for parallelization and faster computation. The temporal convolutional block consists of 1-D causal convolutions followed by a Gated Linear Unit (GLU) to introduce non-linearity and control the flow of information.

For each node in the graph \( G \), the temporal convolution layer explores \( K_t \) neighboring elements along the time axis. This approach does not require padding, and as a result, the length of the sequence decreases by \( K_t-1 \) at each layer. Given an input sequence \( Y \in \mathbb{R}^{M \times C_i} \) with \( M \) time steps and \( C_i \) channels, the convolution kernel \( \Gamma \in \mathbb{R}^{K_t \times C_i \times 2C_o} \) maps the input to a single output element \( [P \ Q] \in \mathbb{R}^{(M-K_t+1) \times (2C_o)} \). The GLU is then applied, splitting \( [P \ Q] \) into two parts, and the output of the temporal convolution is given by:

\[
\Gamma *_T Y = P \odot \sigma(Q) \in \mathbb{R}^{(M-K_t+1) \times C_o},
\]

where \( P \) and \( Q \) are the inputs of the GLU, \( \odot \) denotes the element-wise Hadamard product, and \( \sigma \) is the sigmoid function. The GLU selectively gates the information flow, determining which parts of the input sequence are relevant for capturing dynamic temporal dependencies. Stacking multiple layers of these temporal convolutions enables the model to capture both short- and long-term patterns effectively.

This approach can also be generalized to 3D tensors, where the same convolution kernel is applied to every node \( Y_i \in \mathbb{R}^{M \times C_i} \) in the graph \( G \), resulting in the operation \( \Gamma *_T Y \) with \( Y \in \mathbb{R}^{M \times n \times C_i} \).

\subsection{Dimensionality Reduction/Expansion Module}

\begin{figure}[!hbt]
    \centering
    \includegraphics[width=1\linewidth]{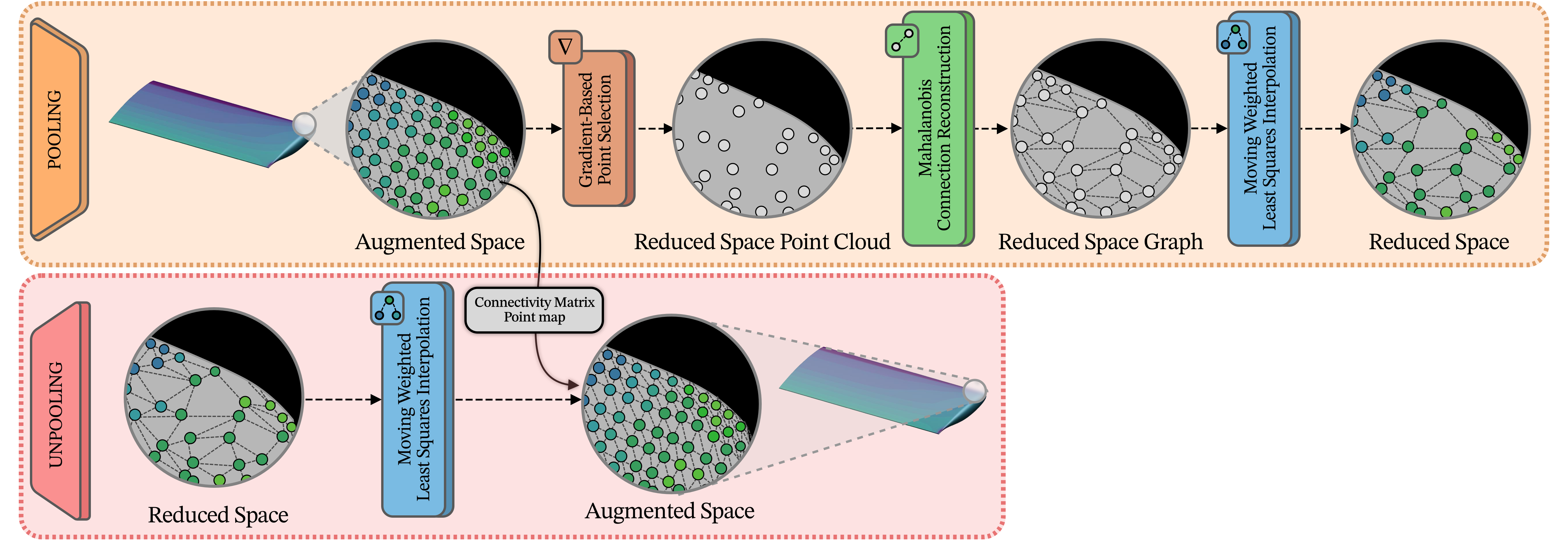}
    \caption{Diagram of the pooling and unpooling modules used in the AE for dimensionality reduction and reconstruction.}
    \label{fig:dimredexp}
\end{figure}

The dimensionality reduction and expansion process aims to simplify the computational load by eliminating redundant information and concentrating on key regions where nonlinear phenomena occur. This method is based on our previous work~\cite{immordino2024predicting} and is visualized in Figure~\ref{fig:dimredexp}, which illustrates both the pooling (reduction) and unpooling (expansion) phases. These phases form the core of the encoding and decoding operations in the AE architecture.

During the pooling phase, points are selected based on pressure gradients to create a reduced point cloud. This strategy ensures that key regions with high gradients are retained, while areas with lower gradients are simplified. The pressure gradient at each node \(i\) is calculated assuming that pressure \(p\) varies linearly in all spatial directions, as described by:

\begin{equation}
    p_i - p_0 = \Delta p_i = \Delta x_i p_x + \Delta y_i p_y + \Delta z_i p_z
\end{equation}

Here, \(p_0\) represents the pressure at a reference node, while \(\Delta x_i\), \(\Delta y_i\), and \(\Delta z_i\) are the spatial differences between neighboring nodes \(i\). Using a least-squares method, the gradient at each node is determined. Nodes are then selected for the reduced space based on a probability function driven by gradient magnitudes:

\begin{equation}
    p(i) = 1 + \frac{1 - e^{-2i/n}}{1 - e^{-2}} (p_1 - p_n) + p_1 \quad \text{for} \quad i = 1, \ldots, n
\end{equation}

where \(i\) refers to the node index ordered by gradient values, \(n\) is the total number of nodes, and \(p_1\) and \(p_n\) are the probabilities assigned to the highest and lowest gradient values, respectively.

To reconnect the reduced point cloud, Mahalanobis distance is used~\cite{de2000mahalanobis}, which accounts for the spread and covariance of the point distribution. This method helps maintain accurate connectivity in the reduced graph, avoiding false connections caused by proximity errors under Euclidean distance~\cite{immordino2024predicting}. The Mahalanobis distance between two points \(x\) and \(y\) is given by:

\begin{equation}
    D_M(x,y)=\sqrt{(x-y)^T S^{-1}(x-y)}
\end{equation}

where \(S\) is the covariance matrix of the original distribution. 

Once the reduced graph is constructed, node values are interpolated using the Moving Weighted Least Squares (MWLS) method. The interpolation matrix \(I_{S_s \rightarrow S_d}\) maps values from the source grid \(S_s\) to the destination grid \(S_d\). The weights are assigned based on proximity, using the function \( w(\mathbf{x}_i) = e^{-{\|\mathbf{x} - \mathbf{x}_i\|_2}} \). The final interpolated value at each node \(\mathbf{x}_j\) is then computed as:

\begin{equation}
    u(\mathbf{x}_j) = \boldsymbol{\Phi}(\mathbf{x}_j) \mathbf{y}_{S_s}
\end{equation}

where:

\begin{equation}
    \boldsymbol{\Phi}(\mathbf{x}_j) = \mathbf{p}^T(\mathbf{x}_j) (\mathbf{P}^T\mathbf{W}\mathbf{P})^{-1}\mathbf{P}^T\mathbf{W}
\end{equation}

In this equation, \(\mathbf{P}\) represents the design matrix for the source nodes, and \(\mathbf{W}\) is a diagonal matrix containing the Gaussian weights. The interpolation matrix \(I_{S_s \rightarrow S_d}\) is used for pooling, and since it is not invertible, the inverse interpolation matrix \(I_{S_d \rightarrow S_s}\) is computed separately for use in the unpooling phase (decoder).

For a detailed explanation of the entire encoding and decoding process, refer to the work of Immordino et al.~\cite{immordino2024predicting}.

\subsection{Pre-trained Autoencoder}

Our proposed framework leverages an AE architecture, pre-trained for subsequent integration into the complete model. The pre-training phase involved using \( C_P \) data as both input and output to the AE, ensuring the model accurately captures the essential features of the pressure distribution over the wing surface. The training dataset comprised the four signals detailed in Table~\ref{tab_signals_parameters}.

\begin{figure}[!hbt] 
    \centering
    \includegraphics[width=0.99\textwidth]{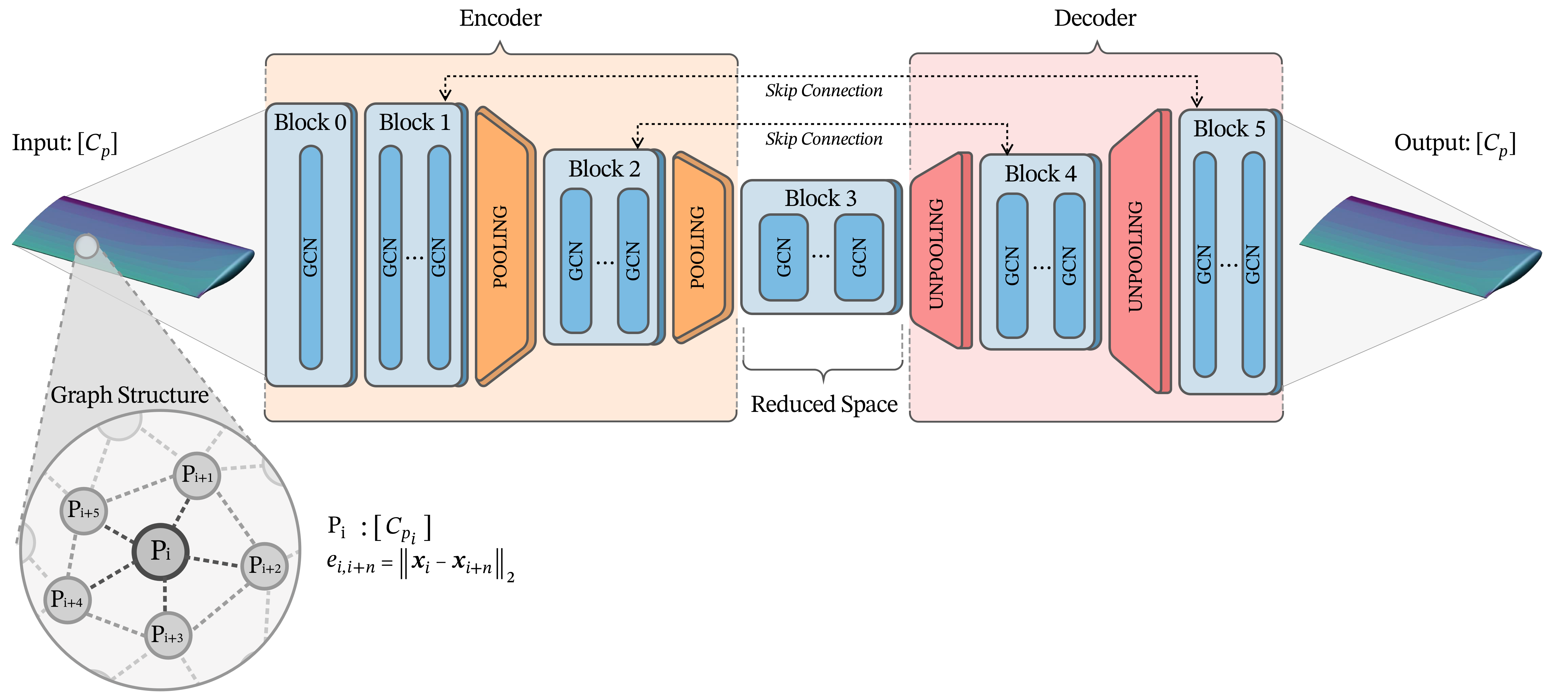}
    \caption{Schematic of the pre-trained AE architecture for compressing and reconstructing the \( C_P \) data within the GST GraphNet framework.}
    \label{fig:pretrained_ae}
\end{figure}

To enhance the robustness of the AE, a data augmentation technique was employed. Specifically, the dataset was augmented by 30\% through the addition of Gaussian noise with 10\% standard deviation of the input data. This augmentation strategy was designed to improve the model ability to generalize and handle variability in the pressure distribution data. Skip connections were integrated before each encoding module to facilitate the direct flow of information across the network. These connections allow the model to bypass certain layers, enabling the retention of critical features and mitigating the risk of information loss during the encoding and decoding processes. The network architecture has been optimized using a Bayesian optimization strategy, following the same approach of our previous work~\cite{immordino2024predicting}. A schematic of the pre-trained AE architecture is shown in Figure~\ref{fig:pretrained_ae}.

\section{Test Case}\label{sec:test_case}

The selected case study for evaluating our framework is the Benchmark Super Critical Wing (BSCW), a transonic rigid semi-span wing featuring a rectangular planform and a supercritical airfoil profile, as detailed in the AIAA Aeroelastic Prediction Workshop~\cite{heeg2013overview}. The freestream conditions for this case are defined by a Mach number of 0.74, a Reynolds number of \( 4.49 \times 10^6 \), and an initial angle of attack of 0 degree. The wing features a reference chord length of 0.4064 $m$ and a surface area of 0.3303 $m^2$, with pitching motion occurring around 30\% of the chord. It is mounted on a flexible support system that allows two degrees of freedom: pitch $\theta$ and plunge $\xi$. It is designed for flutter analysis, presenting challenges due to shock wave motion, shock-induced boundary-layer separation, and the interaction between the shock wave and the detached boundary layer. These nonlinear phenomena pose significant challenges for the framework predictions.

\begin{figure} [!htb] 
    \centering
    \includegraphics[trim=1 1 1 1, clip, width=0.68\textwidth]{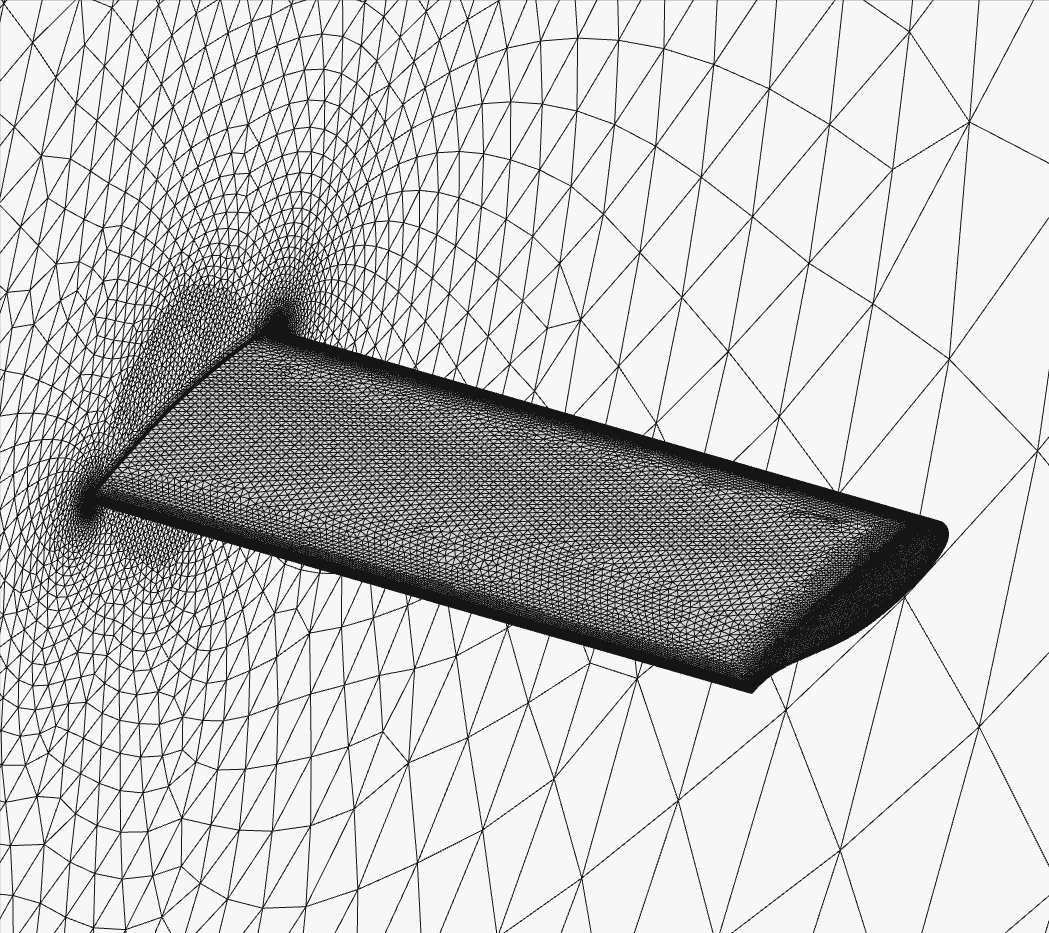}
    \caption{Impression of the BSCW CFD grid.}
    \label{fig:bscw_mesh}
\end{figure}

An unstructured mesh with $8.4 \times 10^6$ elements and 86,840 surface elements was created. A $y^+ = 1$ value was used, based on a preliminary mesh convergence study that confirmed adequate resolution of the boundary layer and shock wave. The computational domain extends 100 chord lengths from the solid wall to the farfield. Figure~\ref{fig:bscw_mesh} provides an impression of the mesh configuration.

The dataset used to train the model was generated with CFD unsteady responses using the Unsteady Reynolds-averaged Navier–Stokes (URANS) formulation with SU2 v7.5.1~\cite{SU2}. The simulations employed the one-equation Spalart-Allmaras turbulence model for RANS closure. To accelerate convergence, a $1v$ multigrid scheme was used. The JST central scheme with artificial dissipation handled convective flow discretization, and the gradients of flow variables were calculated using the Green-Gauss method. The biconjugate gradient stabilization linear solver, along with the ILU preconditioner, was utilized. All URANS simulations started from a steady-state solution, with a timestep of $2 \times 10^{-4}$ seconds and a total simulation time of 2 seconds. These values were chosen to ensure a high temporal resolution, capturing rapid aerodynamic variations while keeping the computational cost manageable over the full simulation period. However, due to computational constraints, the timestep was reduced to $2 \times 10^{-3}$ seconds through downsampling for model training, ensuring a balance between computational feasibility and quality of aerodynamic data.

\begin{figure} [!htb] 
    \centering
    \includegraphics[trim=1 1 1 1, clip, width=0.99\textwidth]{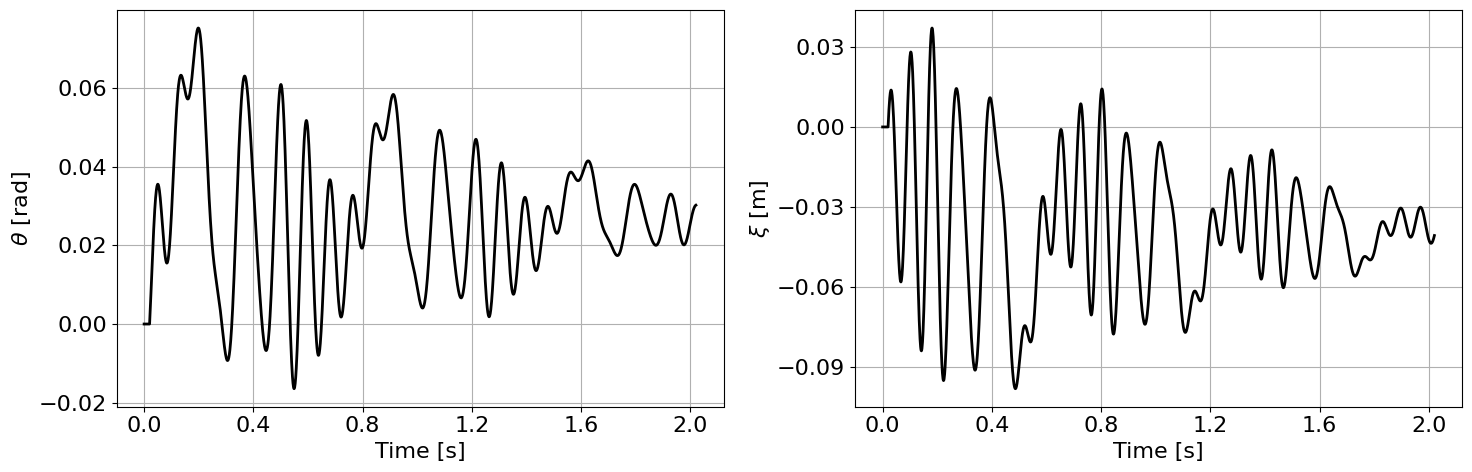}
    \caption{Training signal 1: DS with \(\kappa_\theta = 0.114\), \(a_\theta  = 0.80\) [deg], \(\kappa_\xi = 0.152\), and \(a_\xi  = -0.098\) [m].}
    \label{fig_training_signal1}
\end{figure}

We ran a total of 12 time-varying simulations, divided into training, test, and validation sets as described in Table~\ref{tab_signals_parameters}. The training set comprises damped Schroeder-phased harmonic (DS) simulations (see \ref{app:schroder_signal} for details on the DS signal formulation) with varied combinations of the parameters: reduced frequency for pitch (\(\kappa_\theta\)), maximum pitch amplitude (\(a_\theta)\), reduced frequency for plunge (\(\kappa_\xi\)), and maximum plunge amplitude (\(a_\xi\)), ensuring that the model learns from diverse conditions where these variables exhibit both positive and negative values (Figure~\ref{fig_training_signal1}). This variation helps the model understand the complex interactions between the parameters. The test set extends this diversity by including additional combinations and signal types, specifically undamped Schroeder-phased harmonic (US) and cases focusing solely on \(\theta\) or \(\xi\) variations. This approach allows us to accurately assess the model accuracy and sensitivity in predicting the effects of individual parameters, ensuring that it performs well across different conditions. The use of Schroeder signals is motivated by their ability to effectively cover a broad frequency spectrum while minimizing peak amplitudes, which enhances the model robustness and ability to generalize. The validation set is designed to evaluate the model generalizability to new and unseen data. It includes a scenario similar to those in the training set but with distinct parameter values, as well as a unique single harmonic (SH) signal type, which the model did not encounter during training. This ensures that the model can handle both familiar and novel situations effectively. 
% By structuring the dataset in this way, we ensure comprehensive coverage of possible scenarios, leading to a robust, reliable, and generalizable forecasting model. 

% Schroeder signals are employed in the training process due to their unique ability to cover a broad spectrum of frequencies, which is crucial for enhancing model robustness and generalizability. These signals are constructed by summing multiple sinusoidal components with carefully chosen phases to minimize peak amplitude, leading to an even distribution of energy across the frequency spectrum. The use of nine different phases ensures minimal constructive interference, resulting in a flat spectrum where energy is uniformly spread across a wide range of frequencies. This characteristic is particularly advantageous for training purposes as it exposes the model to a comprehensive range of frequency components present in aerodynamic phenomena. Consequently, the model can better recognize and predict responses to various frequency interactions, ensuring it performs well under diverse and complex conditions. The broad frequency coverage achieved with nine phases also helps avoid overfitting, ensuring that the model remains generalizable and accurate when applied to unseen data.

\begin{table}[ht!]
\centering
\scalebox{0.9}{ % Scale the table to 0.8 of its original size
\begin{tabular}{l c c c c l}
\hline 
\hline 
\textbf{Signals} & \textbf{\boldmath$\kappa_\theta$ [-]} & \textbf{\boldmath$a_\theta $ [deg]} & \textbf{\boldmath$\kappa_\xi$ [-]} & \textbf{\boldmath$a_\xi$ [m]} & \textbf{Type} \\ 
\hline
Training 1 & 0.114 & 0.80 & 0.152 & -0.098 & DS, $\theta$\textgreater0, $\xi$\textless0 \\
Training 2 & 0.114 & -0.80 & 0.152 & 0.098 & DS, $\theta$\textless0, $\xi$\textgreater0 \\
Training 3 & 0.148 & 1.00 & 0.181 & -0.123 & DS, $\theta$\textgreater0, $\xi$\textless0 \\
Training 4 & 0.148 & -1.00 & 0.181 & 0.123 & DS, $\theta$\textless0, $\xi$\textgreater0 \\
\hline
Test 1 & 0.091 & 0.70 & 0.123 & 0.074 & DS, $\theta$\textgreater0, $\xi$\textgreater0 \\
Test 2 & 0.104 & 0.90 & 0.089 & 0.061 & DS, $\theta$\textgreater0, $\xi$\textless0 \\
Test 3 & 0.104 & -0.90 & 0.089 & -0.061 & DS, $\theta$\textless0, $\xi$\textless0 \\
Test 4 & 0.092 & 0.75 & 0.081 & -0.059 & US, $\theta$\textgreater0, $\xi$\textless0 \\
Test 5 & 0.147 & -1.00 & 0.000 & 0.000 & DS, $\theta$\textless0 \\
Test 6 & 0.000 & 0.00 & 0.072 & 0.049 & DS, $\xi$\textgreater0 \\
\hline
Validation 1 & 0.147 & -1.00 & 0.072 & 0.049 & DS, $\theta$\textless0, $\xi$\textgreater0 \\
Validation 2 & 0.106 & 3.00 & 0.089 & -0.246 & SH, $\theta$\textgreater0, $\xi$\textless0 \\
\hline
\hline 
\end{tabular}
}
\caption{Parameters and types of training, test and validation signals. DS: damped Schroeder-phased harmonic, US: undamped Schroeder-phased harmonic, SH: single harmonic.}
\label{tab_signals_parameters}
\end{table}

\section{Results}\label{sec:results}

In this section, we present the results of the reconstructed validation signals for two different types of architectures: feedforward model and ARMAX model. By comparing the performance of these two architectures, we aim to illustrate the influence of incorporating predicted \( C_P \) into the model input on the overall prediction accuracy and robustness. The impact of the temporal layer on the accuracy of each architecture is also studied. 

% The results will highlight the strengths and weaknesses of each approach in predicting unsteady pressure distributions over the wing surface, providing valuable insights into their applicability for different types of aerodynamic analyses.

\subsection{Feedforward Model}

The feedforward model utilizes spatial and temporal data from previous timesteps without relying on its own past predictions, thus preventing the accumulation of errors over time. The model performance across different temporal layers is evaluated using both DS and SH validation signals.

\begin{figure}[!b] \centering \includegraphics[trim=0.2cm 0.2cm 0.2cm 0.2cm, clip, width=1\linewidth]{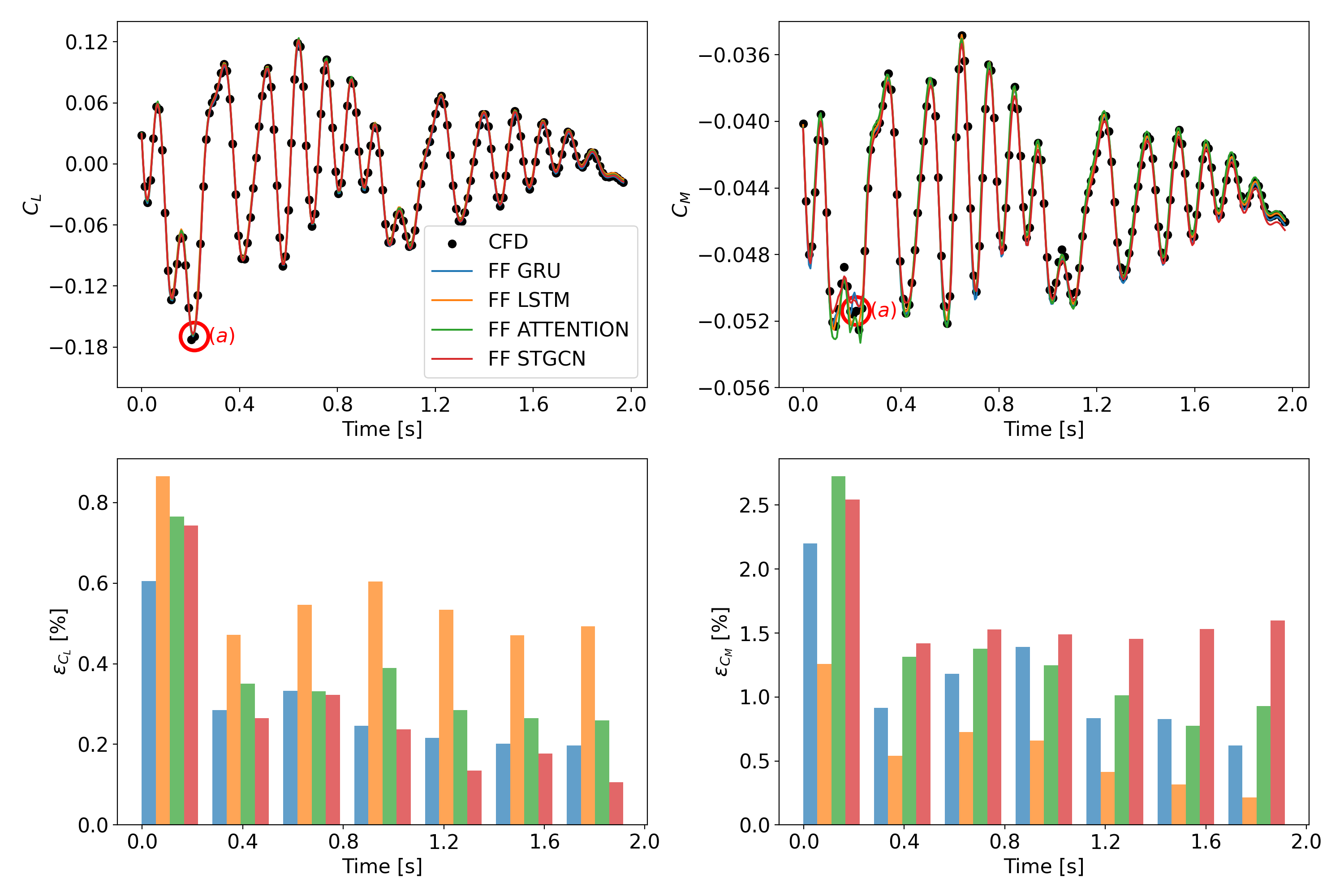} \caption{Validation signal 1 - DS type: Effect of temporal layer selection on $C_L$ and $C_M$ predictions in the feedforward model. The red circle marks the point of maximum error, used for plotting the \( C_P \) distribution. FF: FeedForward.} \label{fig:feedforward_schroeder_plot_CL_CM} \end{figure}

\begin{figure}[!t] \centering \includegraphics[trim=0.2cm 0.2cm 0.2cm 0.2cm, clip, width=1\linewidth]{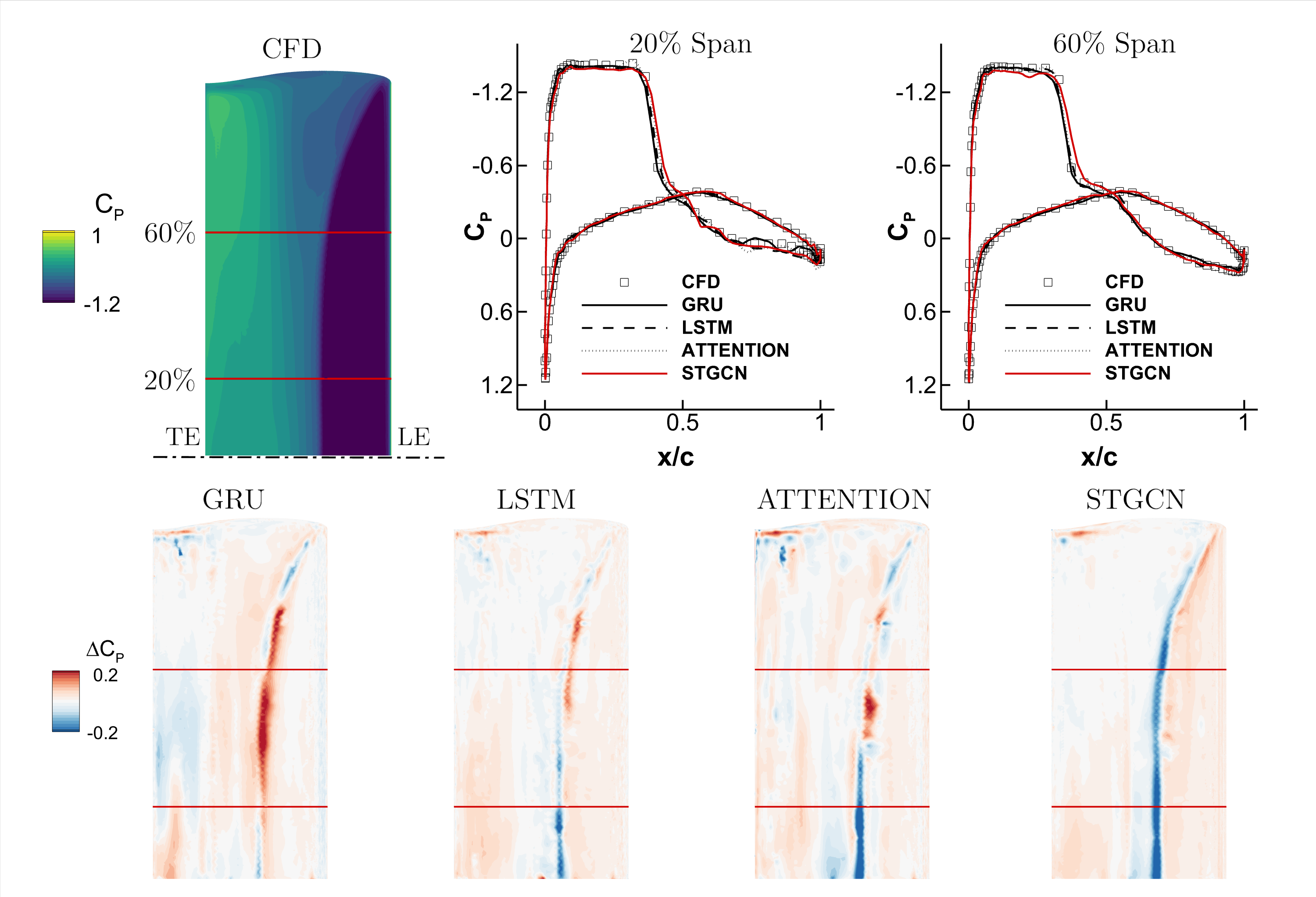} \caption{Validation signal 1 - DS type: Effect of temporal layer selection on $C_P$ prediction at the maximum error point in the feedforward model. The lower surface of the wing is shown. LE: Leading Edge. TE: Trailing Edge. Dash-dot line indicates the symmetry plane.} \label{fig:feedforward_schroeder_plot_CP} \end{figure}

For the DS signal (Figures~\ref{fig:feedforward_schroeder_plot_CL_CM} and \ref{fig:feedforward_schroeder_plot_CP}), the predictions of the aerodynamic coefficients $C_L$ and $C_M$ align well with the CFD reference data in all temporal layers, as shown in Figure~\ref{fig:feedforward_schroeder_plot_CL_CM}. The STGCN layer demonstrates the lowest average error for $C_L$, while LSTM performs slightly better for $C_M$. A region of maximum error, indicated by the red circle, is consistent across all models and represents the point where the predictions for $C_L$ and $C_M$ show the greatest deviation from the reference data. This error appears to be related to the models difficulty in capturing sharp transitions or non-linearities in the aerodynamic flow, particularly near shock waves. Figure~\ref{fig:feedforward_schroeder_plot_CP} shows the \( C_P \) distribution at this maximum error point and highlights the models overall ability to predict the pressure distribution across the wing. While most temporal layers perform reasonably well, some discrepancies near the leading edge and areas affected by shock waves and flow separation are noticeable. These areas, characterized by strong flow gradients and non-linear aerodynamic behavior, are challenging for all models, although STGCN and LSTM exhibit slightly better accuracy compared to GRU and Attention mechanisms.

\begin{figure}[!t] \centering \includegraphics[trim=0.2cm 0.2cm 0.2cm 0.2cm, clip, width=1\linewidth]{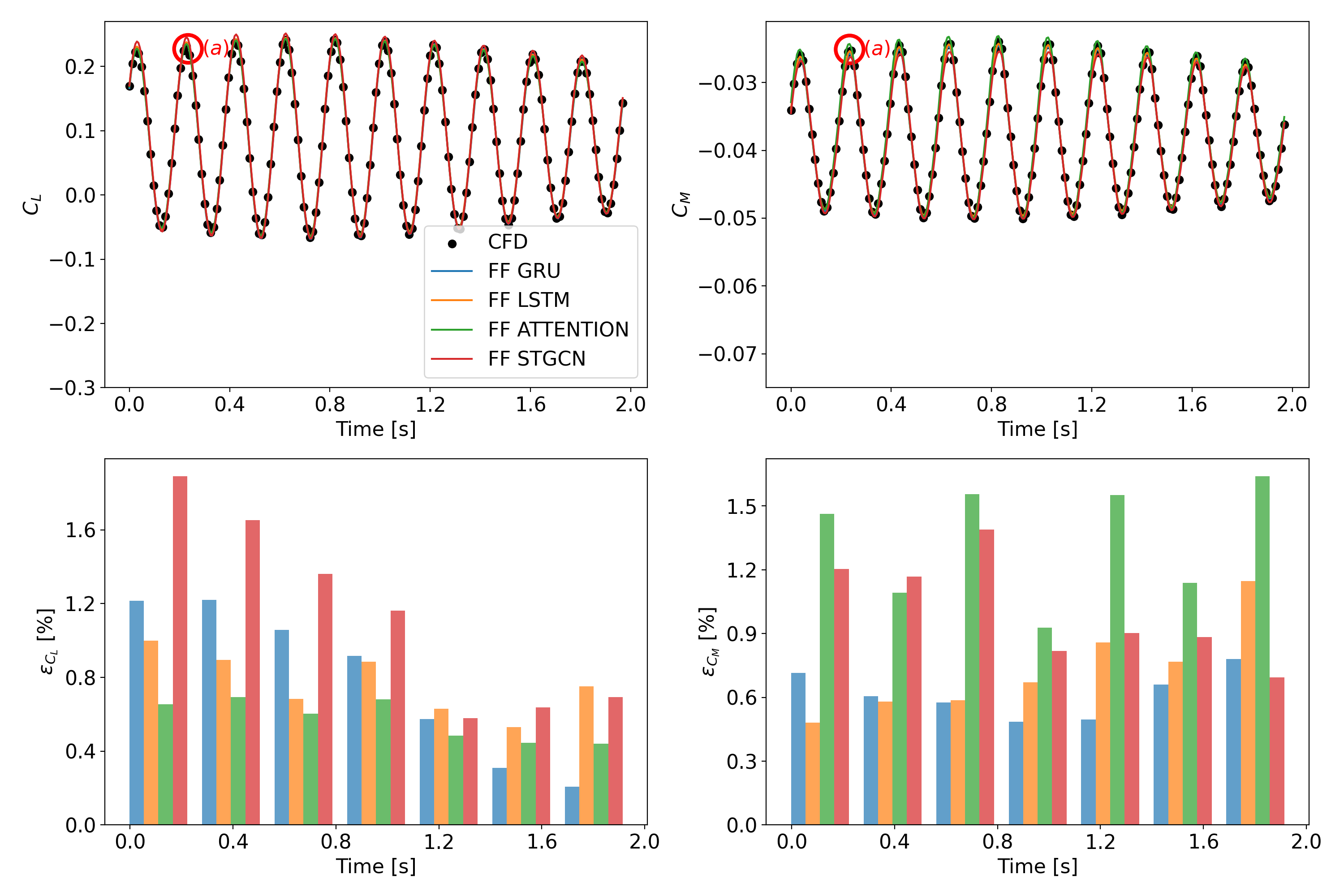} \caption{Validation signal 2 - SH type: Effect of temporal layer selection on $C_L$ and $C_M$ predictions in the feedforward model. The red circle marks the point of maximum error, used for plotting the \( C_P \) distribution. FF: FeedForward} \label{fig:feedforward_harmonic_plot_CL_CM} \end{figure}

\begin{figure}[!b] \centering \includegraphics[trim=0.2cm 0.2cm 0.2cm 0.2cm, clip, width=1\linewidth]{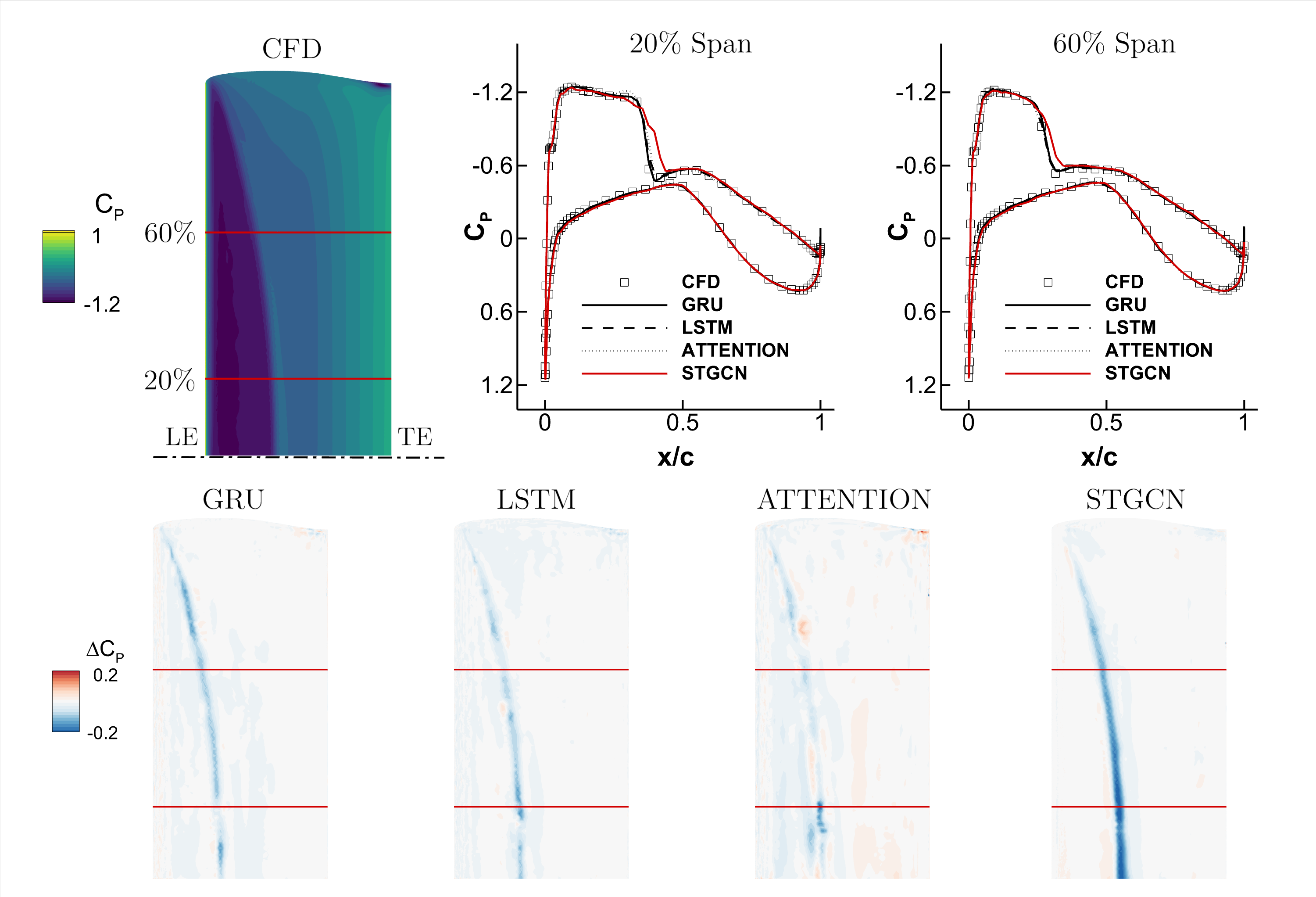} \caption{Validation signal 2 - SH type: Effect of temporal layer selection on $C_P$ prediction at the maximum error point in the feedforward model. The upper surface of the wing is shown. LE: Leading Edge. TE: Trailing Edge. Dash-dot line indicates the symmetry plane.} \label{fig:feedforward_harmonic_plot_CP} \end{figure}

For the SH signal (Figures \ref{fig:feedforward_harmonic_plot_CL_CM} and \ref{fig:feedforward_harmonic_plot_CP}), which involves higher-frequency oscillations, the models encounter increased errors, particularly in predicting peak values for $C_L$ and $C_M$. LSTM and STGCN continue to provide the most accurate results, although both exhibit some phase and amplitude errors due to the rapid oscillations. The red-circled point, indicating the region of maximum error for $C_L$ and $C_M$, again highlights the challenges of accurately capturing high-frequency aerodynamic loads. Figure~\ref{fig:feedforward_harmonic_plot_CP} provides a detailed view of the \( C_P \) distribution at this maximum error point, where the models struggle more to match the rapid fluctuations in $C_P$. In particular, regions near the shock wave, which undergo significant temporal variation, show greater discrepancies. While the general \( C_P \) distribution is captured, the accuracy decreases in areas where shock-induced flow separation occurs, with LSTM and STGCN again showing marginally better performance in these challenging regions.

To quantitatively evaluate the models, we calculated three error metrics—Mean Absolute Percentage Error (\texttt{MAPE}), coefficient of determination (\texttt{R2}), and Root Mean Square Error (\texttt{RMSE})—for \(C_P\) predictions across both validation signals, as shown in Table \ref{tab:feedforward_signal_comparison}. The \texttt{MAPE} was derived by averaging the absolute errors across each timestep, weighted by the corresponding cell area, and normalized accordingly. It reflects the average percentage error across predictions, showing that the LSTM model consistently achieves the lowest values, while the STGCN model follows closely. GRU and Attention mechanisms, on the other hand, display significantly higher \texttt{MAPE}, particularly for the high-frequency oscillations of the SH signal, indicating their difficulty in capturing rapid temporal variations and non-linearities. In terms of \texttt{R2}, which measures how well the model explains variance in the data, LSTM again performs the best, capturing the highest degree of variability, followed closely by STGCN, which also demonstrates strong performance in the DS signal but slightly lower accuracy for the SH signal. GRU and Attention layers show lower \texttt{R2}, further confirming their difficulty to fully capture the complexities in regions involving shock waves and high dynamic variability. Regarding \texttt{RMSE}, which emphasizes larger errors due to its quadratic nature, the LSTM model continues to perform better with the lowest values, indicating a good robustness against significant deviations. STGCN performs similarly but struggles slightly more with large deviations in dynamic conditions like the SH signal. Again, GRU and Attention mechanisms exhibit the highest \texttt{RMSE}. These results suggest that all models are well-suited for capturing both spatial and temporal dependencies with good overall performance, but GRU and Attention layers struggle more to capture the rapid temporal variations and non-linearities in the \(C_P\) distribution.

In conclusion, while all temporal layers provide reasonable predictions for unsteady aerodynamic phenomena, the LSTM model delivers the most robust performance across both validation signals. STGCN also performs well, particularly for $C_L$ predictions, while GRU and Attention mechanisms exhibit larger errors, especially in more complex dynamics. The evaluation of \( C_P \) distribution highlights the importance of accurately capturing non-linear flow features, with regions near shock waves and flow separations proving to be the most challenging for all models.

\begin{table}[!t]
\centering
\scalebox{0.9}{
\begin{tabular}{l c c c c c c}
\hline
\hline
\textbf{Temporal Layer} & \multicolumn{2}{c}{\textbf{MAPE}} & \multicolumn{2}{c}{\textbf{R2}} & \multicolumn{2}{c}{\textbf{RMSE}} \\
               & DS & SH & DS & SH & DS & SH \\
\hline
GRU        & 1.2382    & 1.7851    & 0.9851    & 0.9830    & 0.0217    & 0.0229 \\
LSTM       & \textbf{0.7471}    & \textbf{0.9695}    & \textbf{0.9937}    & \textbf{0.9909}    & 0.0194    & 0.0215 \\
Attention  & 1.0470    & 1.5452    & 0.9887    & 0.9815    & 0.0238    & 0.0291 \\
STGCN      & 0.8524    & 0.9975    & 0.9918    & 0.9897    & \textbf{0.0163}    & \textbf{0.0181} \\
\hline
\hline
\end{tabular}
}
\caption{Comparison of \texttt{MAPE}, \texttt{R2}, and \texttt{RMSE} for \(C_P\) predictions in the feedforward model with different temporal layers for DS and SH validation signals.}
\label{tab:feedforward_signal_comparison}
\end{table}

\subsection{ARMAX Model}

\begin{figure}[!hbt]
    \centering
    \includegraphics[trim=0.2cm 0.2cm 0.2cm 0.2cm, clip, width=1\linewidth]{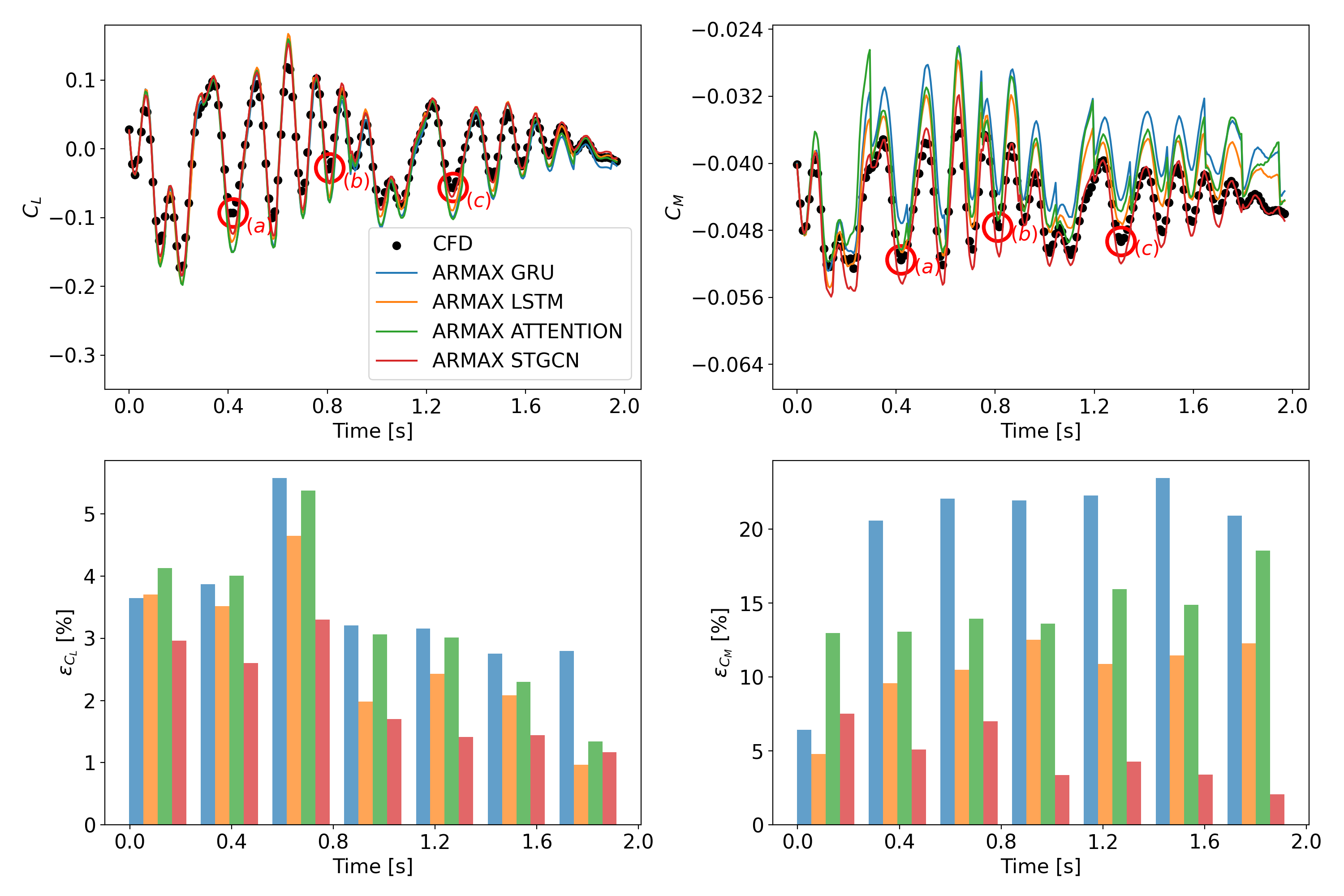}
    \caption{Validation signal 1 - DS type: Impact of temporal layer selection on $C_L$ and $C_M$ predictions in the ARMAX model. Red circles denote the points used for plotting the \( C_P \) distribution.}
    \label{fig:armax_schroeder_plot_CL_CM}
\end{figure}

The ARMAX model, which integrates past predictions into its input, was evaluated using various temporal layers to assess its performance in predicting \(C_L\), \(C_M\), and \(C_P\) across DS and SH validation signals. As shown in Figure~\ref{fig:armax_schroeder_plot_CL_CM}, one of the biggest limitations of the ARMAX model consists of accumulation of errors over time, particularly when using the GRU and Attention mechanisms. Again, LSTM and STGCN demonstrate better predictive performance, although both exhibit some deviations in complex flow regions. The red-circled points were selected to visualize the evolution of the error over time. These points were spaced at regular intervals along the signal to provide insight into how prediction accuracy changes throughout the sequence. This allows us to observe how the model handles different stages of prediction and highlights its difficulty in accurately capturing sharp transitions and non-linearities in the aerodynamic flow, particularly around shocks. This issue is further emphasized in Figure~\ref{fig:armax_schroeder_plot_CP}, where the \(C_P\) at these selected points shows that the ARMAX model has more difficulty maintaining accuracy near the leading edge and in shock-affected regions. Despite these challenges, LSTM and STGCN continue to provide the most reliable predictions despite the inherent error accumulation.

\begin{figure}[!hbt]
    \centering
    \includegraphics[trim=0.2cm 0.2cm 0.2cm 0.2cm, clip, width=0.8\linewidth]{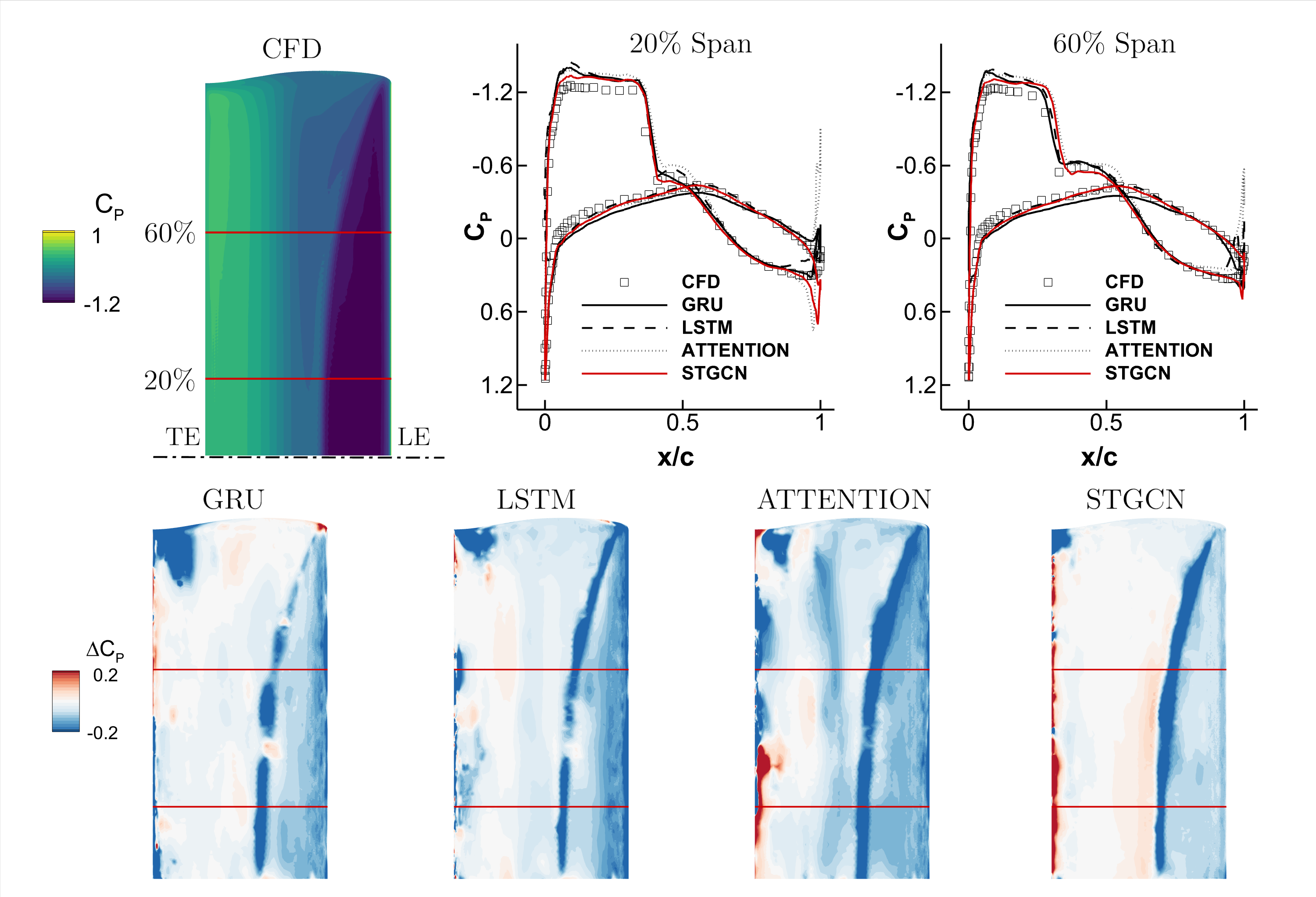}
    \caption*{Time Instance (a)}
\end{figure}
\begin{figure}[!hbt]
    \centering
    \includegraphics[trim=0.2cm 0.2cm 0.2cm 0.2cm, clip, width=0.8\linewidth]{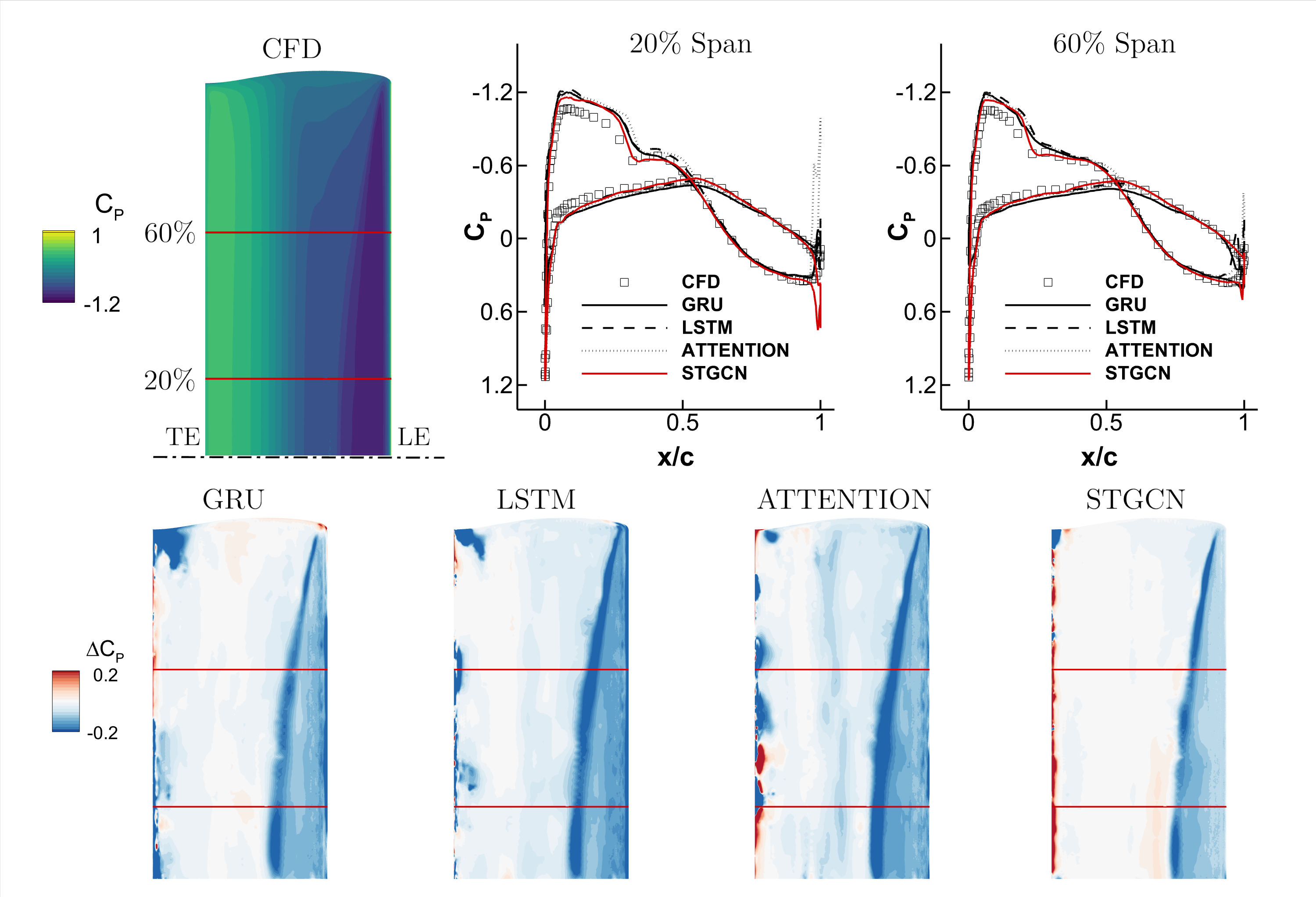}
    \caption*{Time Instance (b)}
\end{figure}
\begin{figure}[!hbt]
    \centering
    \includegraphics[trim=0.2cm 0.2cm 0.2cm 0.2cm, clip, width=0.8\linewidth]{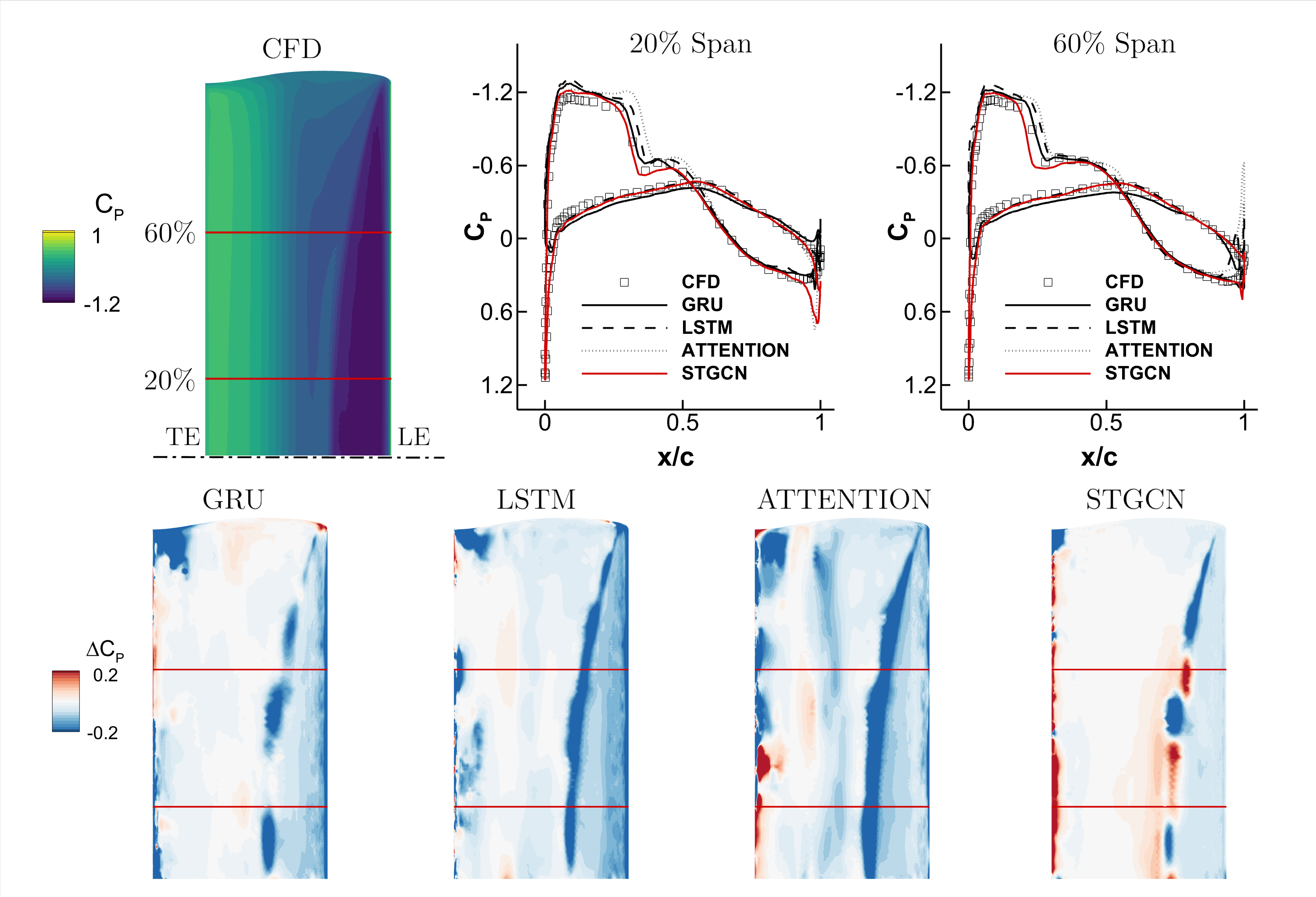}
    \caption*{Time Instance (c)}
\end{figure}
\begin{figure}[!hbt]
    \centering
    \caption{Validation signal 1 - DS type: Impact of temporal layer selection on $C_P$ prediction at the maximum error point in the ARMAX model. The lower surface of the wing is shown. LE: Leading Edge. TE: Trailing Edge. Dash-dot line indicates the symmetry plane.}
    \label{fig:armax_schroeder_plot_CP}
\end{figure}

\begin{figure}[!hbt]
    \centering
    \includegraphics[trim=0.2cm 0.2cm 0.2cm 0.2cm, clip, width=1\linewidth]{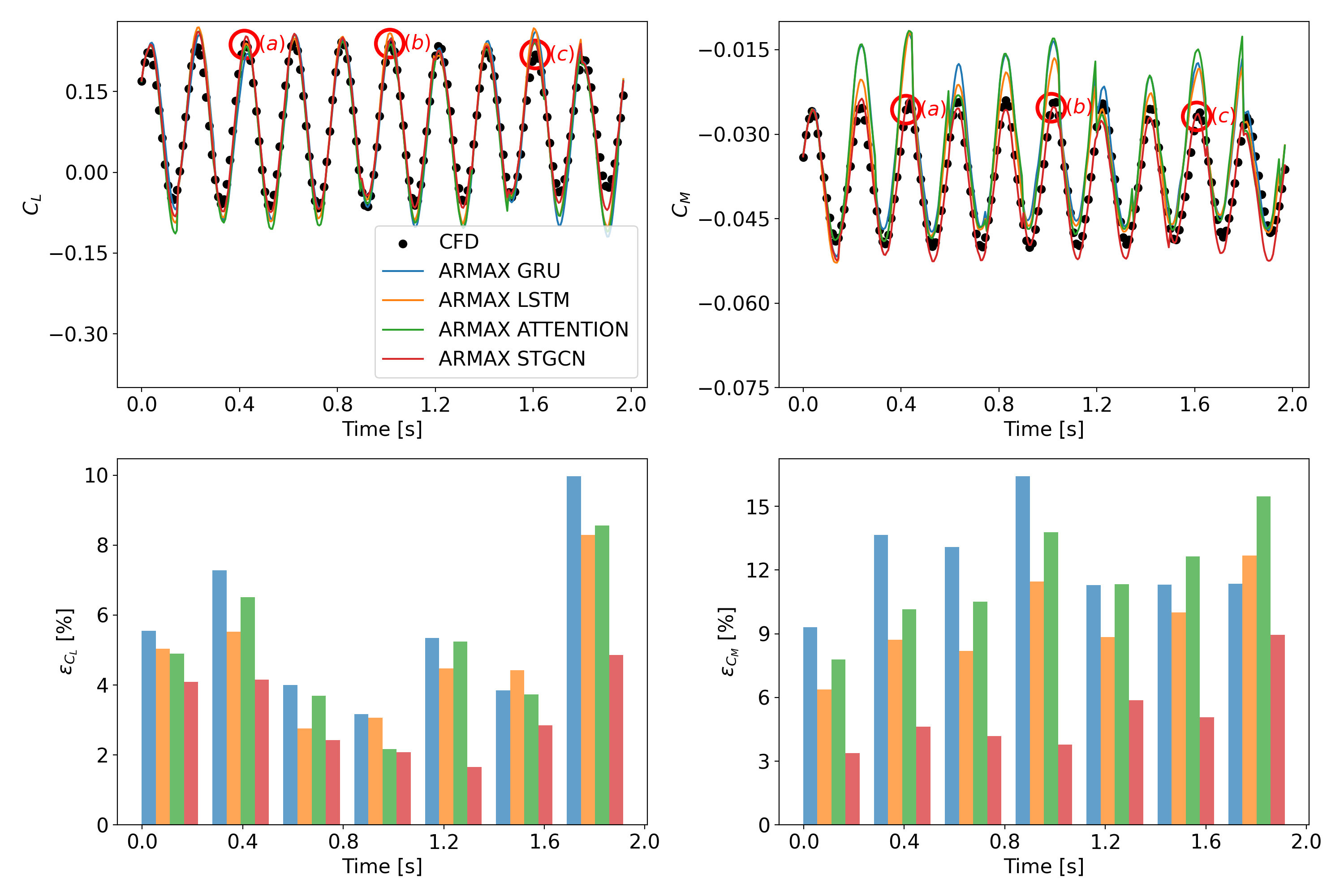}
    \caption{Validation signal 2 - SH type: Impact of temporal layer selection on $C_L$ and $C_M$ predictions in the ARMAX model. Red circles denote the points used for plotting the \( C_P \) distribution.}
    \label{fig:armax_harmonic_plot_CL_CM}
\end{figure}

For the SH signal (Figures \ref{fig:armax_harmonic_plot_CL_CM} and \ref{fig:armax_harmonic_plot_CP}), which involves higher frequency oscillations, the ARMAX model exhibits increased errors, particularly for the GRU and Attention layers. The rapid oscillations introduce phase and amplitude discrepancies, with LSTM and STGCN handling these variations better, though both models still show increased errors compared to the DS signal. The \( C_P \) distribution in Figure~\ref{fig:armax_harmonic_plot_CP} reveals that the ARMAX model is more susceptible to error propagation in highly dynamic regions near shock waves, where rapid changes in flow introduce greater inaccuracies. Although LSTM and STGCN mitigate these issues to some extent, they still experience some degradation in predictive accuracy due to the model autoregressive nature.

The superior performance of LSTM and STGCN can be attributed to their inherent design, which allows them to handle temporal dependencies more effectively. The LSTM architecture, with its complex gating mechanisms, enables the model to retain and manage both long- and short-term dependencies, preventing information loss over multiple timesteps and helping to mitigate the error accumulation issue inherent in the ARMAX model. The STGCN layer combines both spatial and temporal convolutions, making it well-suited to handle non-uniform grid structures and effectively capture spatial correlations (such as pressure gradients across the wing) as well as temporal dependencies during rapid changes in aerodynamic conditions, as seen in the SH signal.

In contrast, the GRU simpler structure limits its capacity to capture long-term dependencies effectively. The Attention mechanism, while effective for capturing important temporal relationships in longer sequences, is not as well suited to the short input sequences used in this model, where the benefits of dynamically weighting timesteps are reduced, limiting the Attention layer effectiveness.

\begin{figure}[!hbt]
    \centering
    \includegraphics[trim=0.2cm 0.2cm 0.2cm 0.2cm, clip, width=0.8\linewidth]{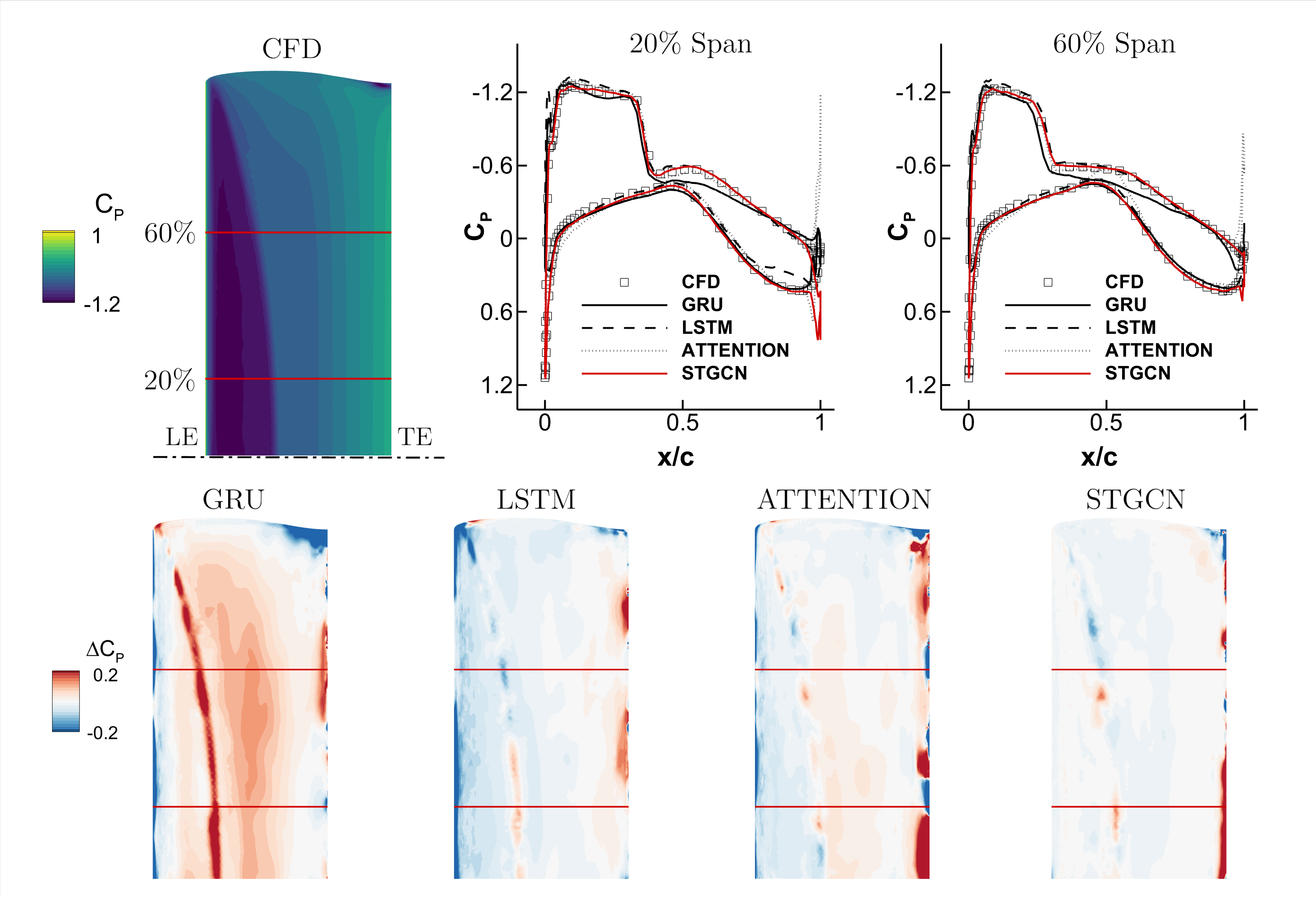}
    \caption*{Time Instance (a)}
\end{figure}
\begin{figure}[!hbt]
    \centering
    \includegraphics[trim=0.2cm 0.2cm 0.2cm 0.2cm, clip, width=0.8\linewidth]{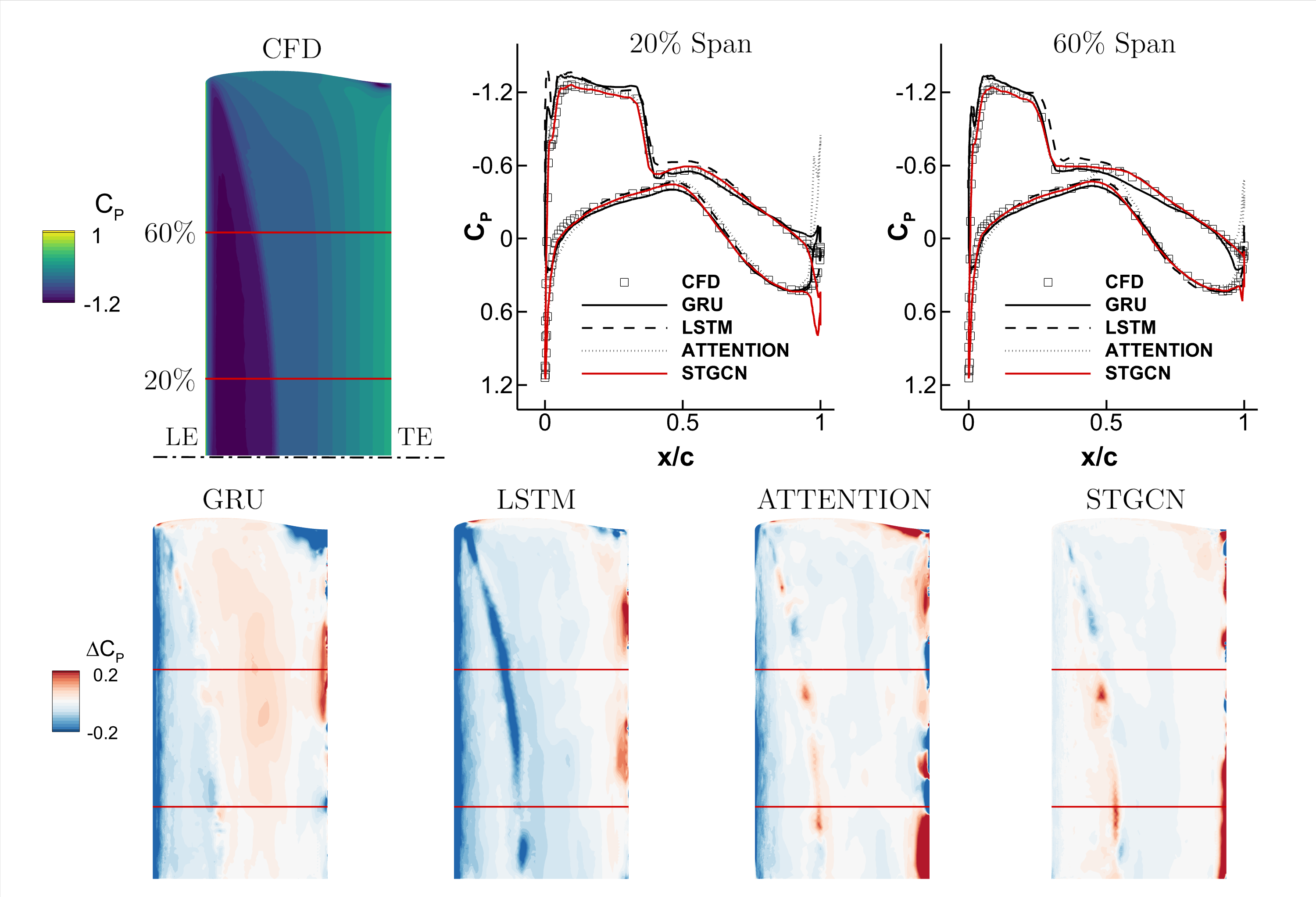}
    \caption*{Time Instance (b)}
\end{figure}
\begin{figure}[!hbt]
    \centering
    \includegraphics[trim=0.2cm 0.2cm 0.2cm 0.2cm, clip, width=0.8\linewidth]{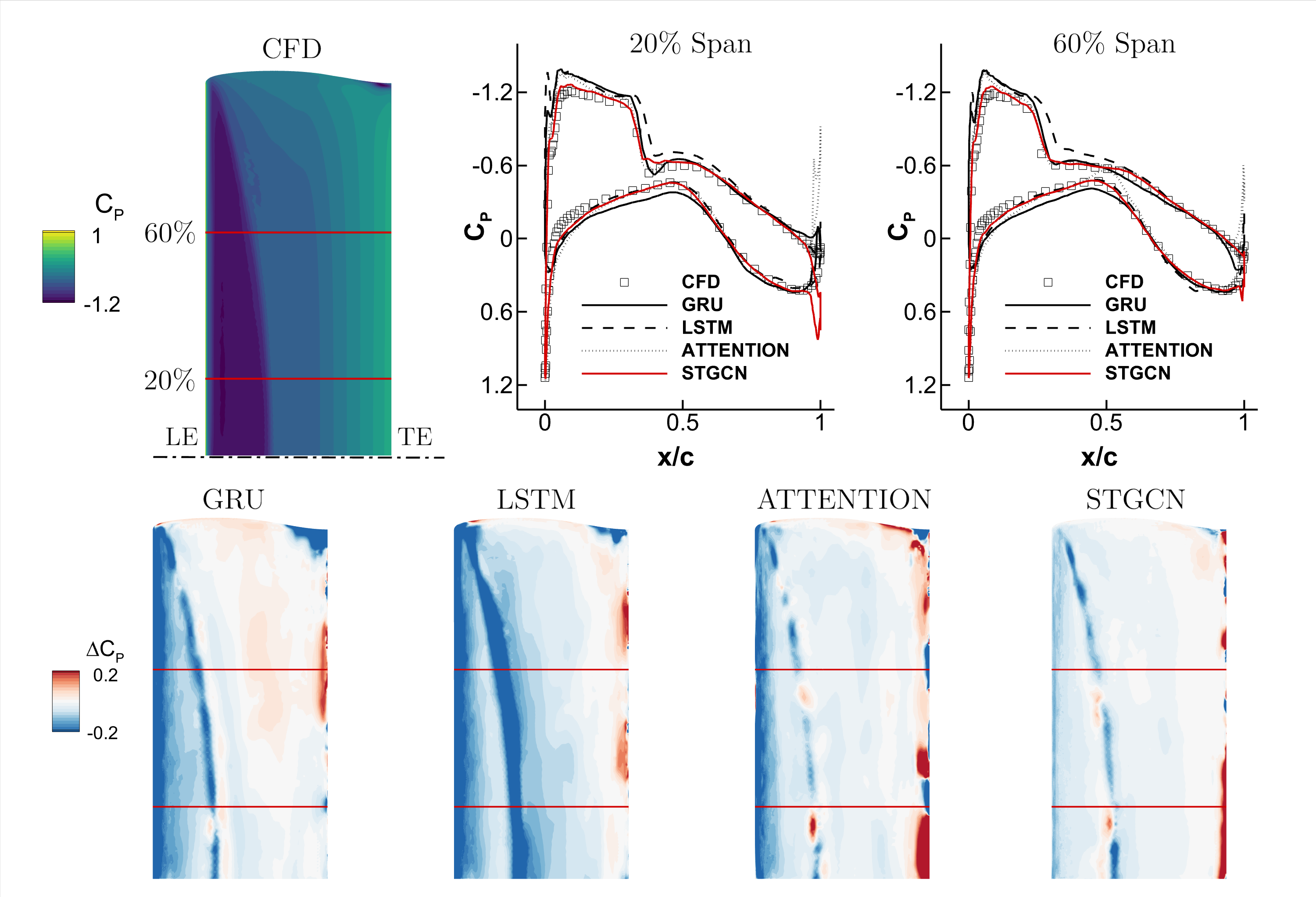}
    \caption*{Time Instance (c)}
\end{figure}
\begin{figure}[!hbt]
    \centering
    \caption{Validation signal 2 - SH type: Impact of temporal layer selection on $C_P$ prediction at the maximum error point in the ARMAX model. The upper surface of the wing is shown. LE: Leading Edge. TE: Trailing Edge. Dash-dot line indicates the symmetry plane.}
    \label{fig:armax_harmonic_plot_CP}
\end{figure}

Table~\ref{tab:armax_signal_comparison} quantifies the performance of the different temporal layers in terms of \texttt{MAPE}, \texttt{R2}, and \texttt{RMSE} for \(C_P\) predictions in the ARMAX model. The STGCN model consistently achieves the lowest \texttt{MAPE} and \texttt{RMSE} values across both the DS and SH signals, with LSTM performing nearly as well. This suggests that both models are adept at handling temporal dependencies even in autoregressive contexts, although the STGCN slightly outperforms LSTM, particularly in the dynamic SH signal. GRU and Attention layers, on the other hand, show higher \texttt{MAPE} and \texttt{RMSE}, along with lower \texttt{R2}, indicating that they struggle to mitigate the error accumulation inherent in the ARMAX model, particularly in high-frequency oscillations. The lower \texttt{R2} values for GRU and Attention mechanisms highlight their reduced ability to capture the full variability in the data with rapid aerodynamic changes. Overall, the results suggest that while the ARMAX model can capture general trends, its susceptibility to error propagation makes it crucial to use robust temporal layers, such as LSTM and STGCN, to minimize predictive inaccuracies in highly dynamic and non-linear aerodynamic conditions.

\begin{table}[!htb]
\centering
\scalebox{0.9}{
\begin{tabular}{l c c c c c c}
\hline
\hline
\textbf{Temporal Layer} & \multicolumn{2}{c}{\textbf{MAPE}} & \multicolumn{2}{c}{\textbf{R2}} & \multicolumn{2}{c}{\textbf{RMSE}} \\
               & DS & SH & DS & SH & DS & SH \\
\hline
GRU        & 8.5577    & 6.7837    & 0.7560    & 0.7993    & 0.1389    & 0.1439 \\
LSTM       & 7.7511    & 6.4299    & 0.8426    & 0.8506    & 0.1002    &  0.0864  \\
Attention  & 7.7985    & 5.8077    & 0.8167    & 0.7796    & 0.1233    & 0.1048 \\
STGCN      & \textbf{6.9381}    & \textbf{5.7971} & \textbf{0.8571}    & \textbf{0.8648}    & \textbf{0.0938}    & \textbf{0.0844}    \\
\hline
\hline
\end{tabular}
}
\caption{Comparison of \texttt{MAPE}, \texttt{R2}, and \texttt{RMSE} for \(C_P\) predictions in the ARMAX model with different temporal layers for DS and SH validation signals.}
\label{tab:armax_signal_comparison}
\end{table}

%%%%%%%%%%%%%%%%%%%%

\subsection{Feedforward vs ARMAX}

\begin{figure}[!hbt]
    \centering
    \includegraphics[trim=0.2cm 0.2cm 0.2cm 0.2cm, clip, width=1\linewidth]{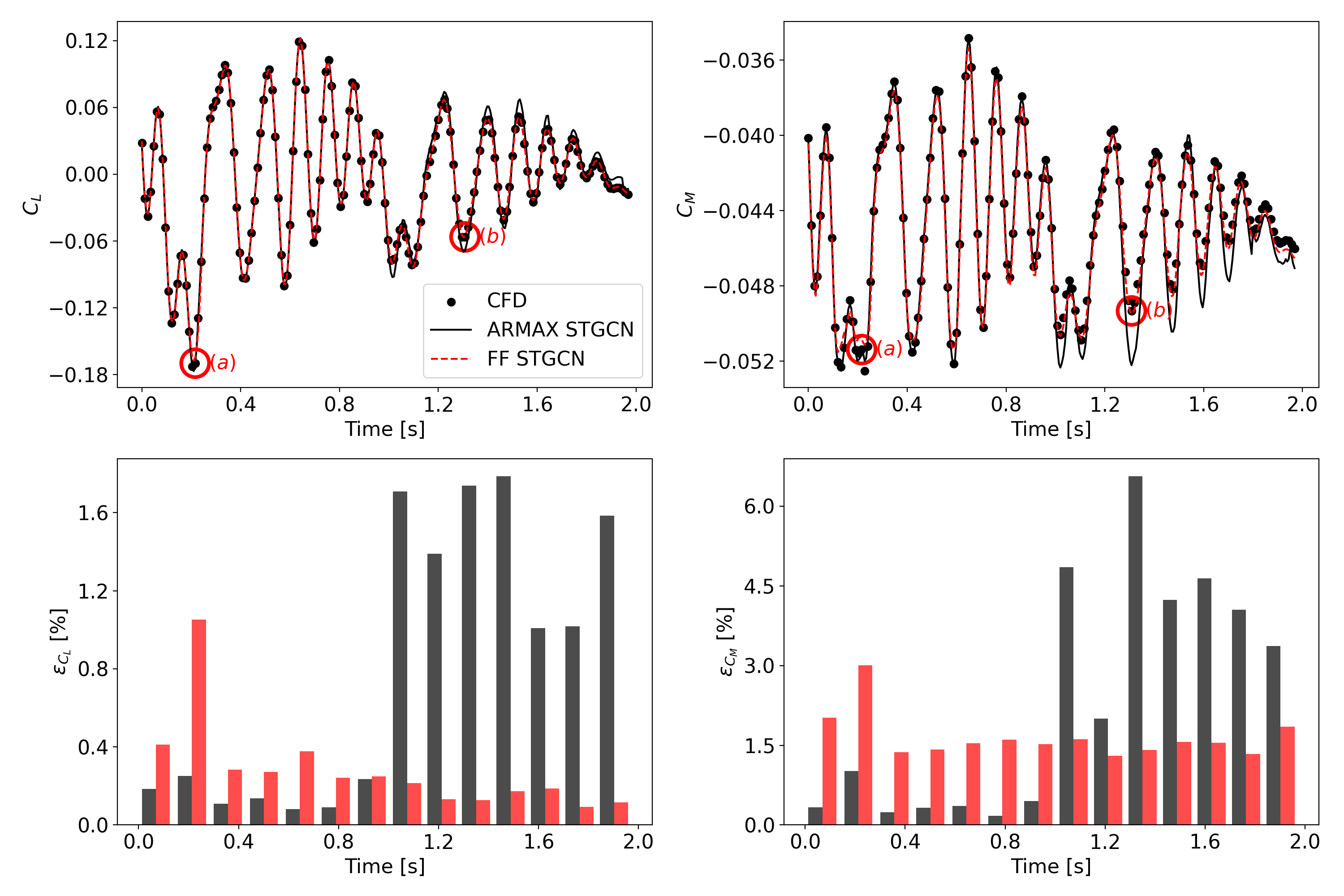}
    \caption{Validation signal 1 - DS type: ARMAX vs Feedforward model using STGCN temporal layer on $C_L$ and $C_M$ predictions. Red circles denote the points used for plotting the \( C_P \) distribution.}
    \label{fig:FFvsARMAX_schroeder_plot_CL_CM}
\end{figure}

\begin{figure}[!hbt]
    \centering
    \includegraphics[trim=0.2cm 0.2cm 0.2cm 0.2cm, clip, width=0.5\linewidth]{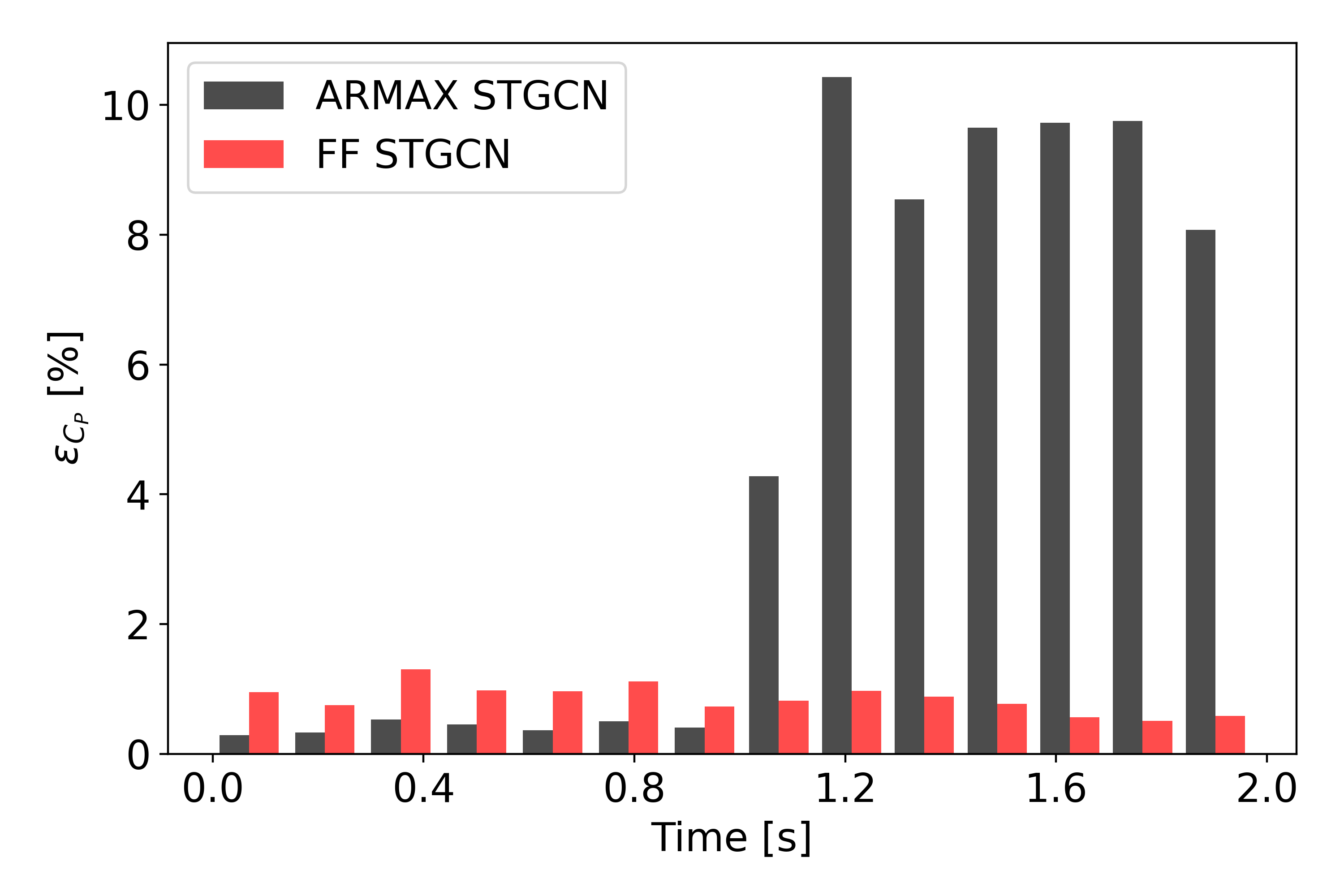}
    \caption{Validation signal 1 - DS type: Evolution of MAPE of $C_P$ with ARMAX and Feedforward model.}
    \label{fig:FFvsARMAX_schroeder_plot_MAPE_CP}
\end{figure}

The comparison between ARMAX and feedforward models, both using the STGCN temporal layer, reveals notable differences in performance, especially regarding error accumulation and prediction stability. The STGCN temporal layer was selected for this comparison because it consistently yielded the most accurate results across both validation signals, as demonstrated in previous sections. In this case, ARMAX model uses ground-truth \(C_P\) values for the first half of the signal, after which it switches to using its own predictions for subsequent timesteps. As shown in Figure~\ref{fig:FFvsARMAX_schroeder_plot_CL_CM}, the feedforward model provides more accurate predictions for $C_L$ and $C_M$ on the DS signal, avoiding the accumulation of errors observed in the ARMAX model. The red circled points were chosen to be in the middle of the first and second halves of the signal, providing information on the behavior of the model during the transition from using ground truth to self-predicted values. This choice allows for a clearer comparison of model performance during both phases, highlighting the ARMAX model difficulty in limiting error accumulation, while the feedforward model maintains closer alignment with the reference data. This trend is further confirmed in Figure~\ref{fig:FFvsARMAX_schroeder_plot_MAPE_CP}, where the evolution of the \texttt{MAPE} in \(C_P\) reveals that the feedforward model consistently maintains lower error levels over time compared to the ARMAX model, which, as expected, shows a clear pattern of accumulation of errors.

Interestingly, the ARMAX model performs relatively well during the first half of the signals, when ground-truth \(C_P\) values are used as input. In this phase, the ARMAX model can even outperform the feedforward model, as seen in the initial part of Figures \ref{fig:FFvsARMAX_schroeder_plot_CL_CM} and \ref{fig:FFvsARMAX_schroeder_plot_MAPE_CP}. However, once the model begins using its own predicted \(C_P\) values for subsequent timesteps, error accumulation begins, resulting in larger deviations from the reference data. This behavior is clearly seen in Figure~\ref{fig:FFvsARMAX_schroeder_plot_CP}, where the ARMAX model shows growing discrepancies in \(C_P\) predictions as the autoregressive process progresses. In contrast, the feedforward model avoids this problem by not relying on its past predictions, allowing it to maintain better accuracy in regions near the leading and trailing edges, where flow complexity is higher.

\begin{figure}[!hbt]
    \centering
    \includegraphics[trim=0.2cm 0.2cm 0.2cm 0.2cm, clip, width=0.8\linewidth]{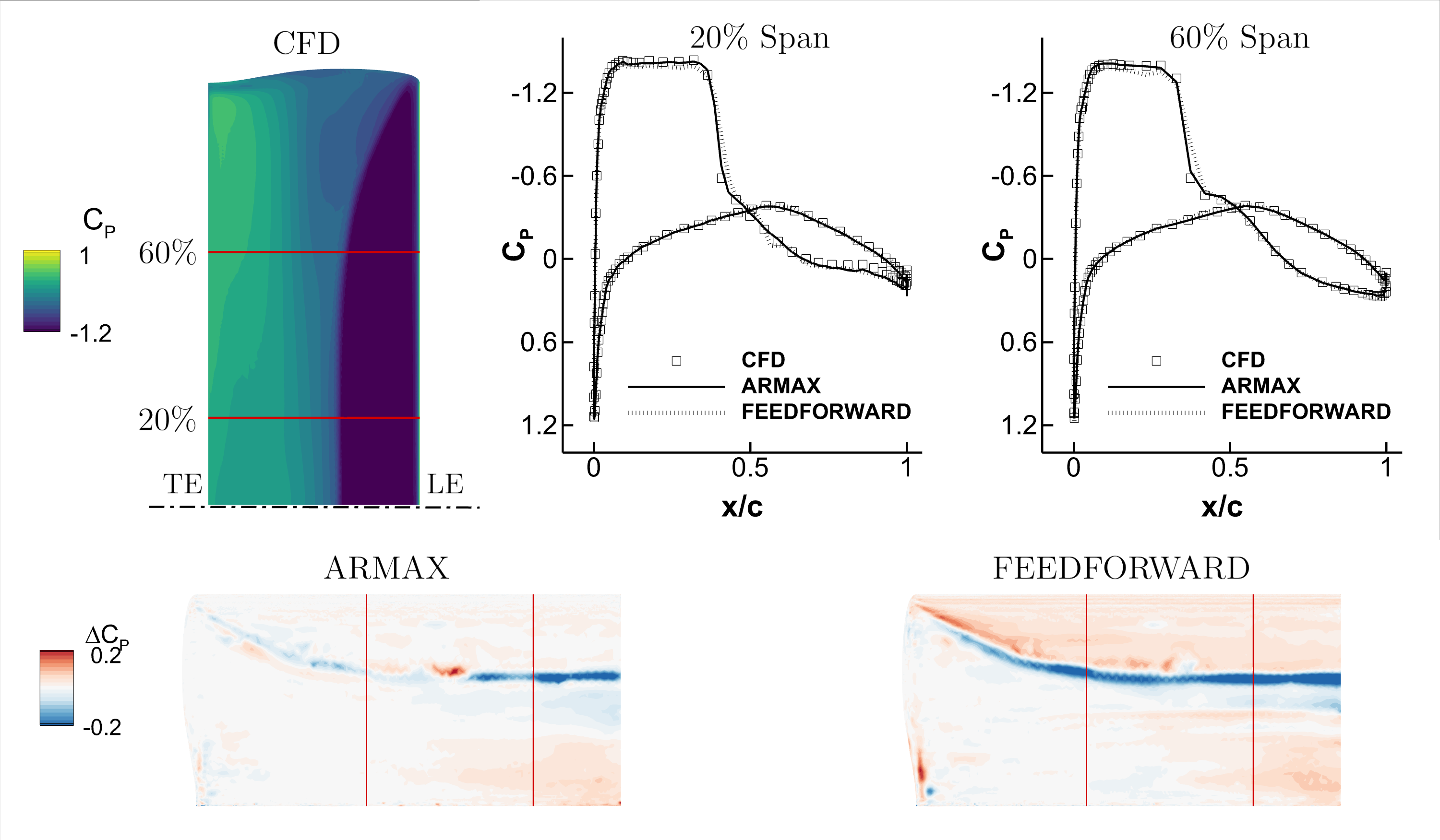}
    \caption*{Time Instance (a)}
\end{figure}
\begin{figure}[!hbt]
    \centering
    \includegraphics[trim=0.2cm 0.2cm 0.2cm 0.2cm, clip, width=0.8\linewidth]{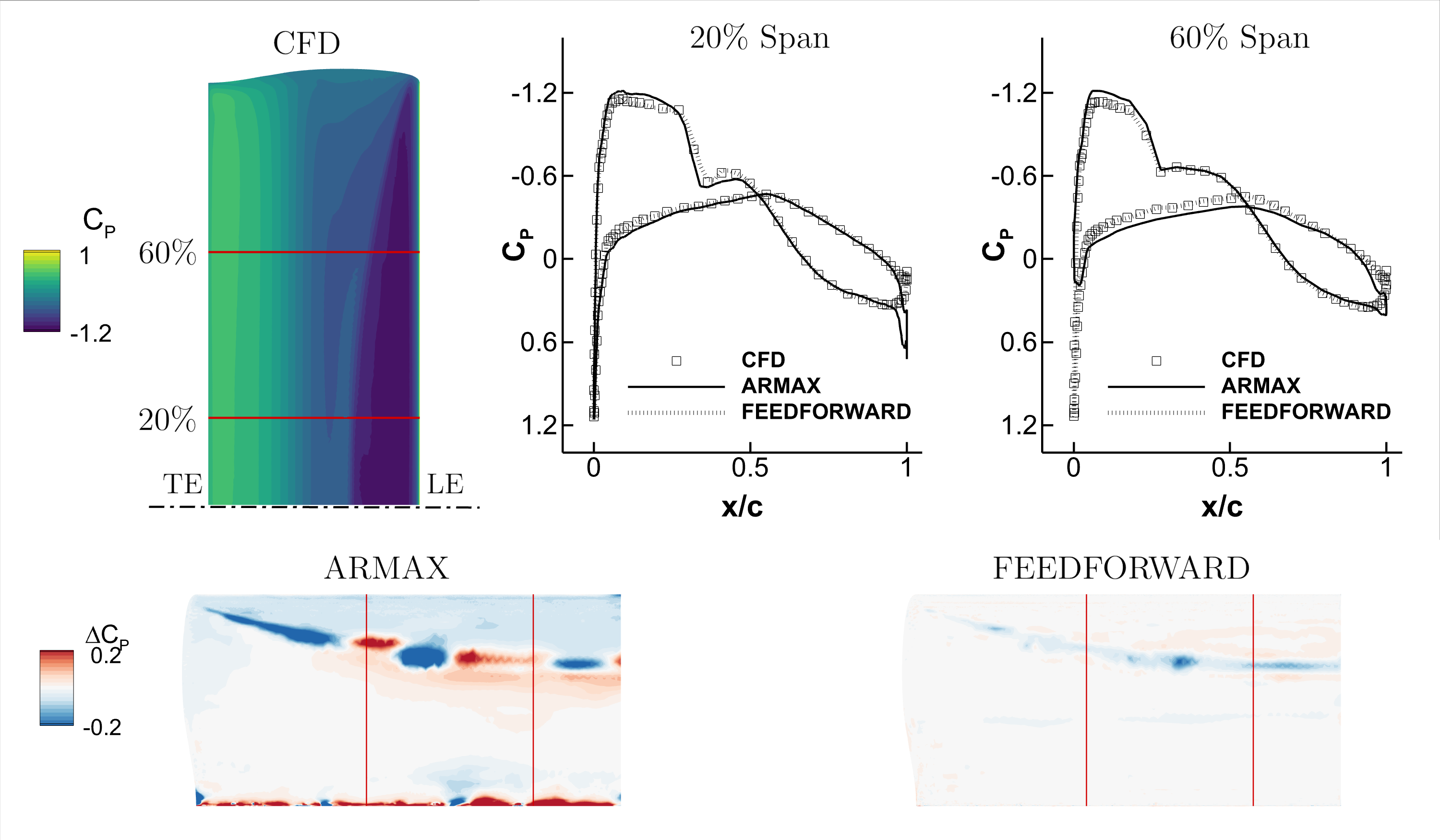}
    \caption*{Time Instance (b)}
\end{figure}
\begin{figure}[!hbt]
    \centering
    \caption{Validation signal 1 - DS type: ARMAX vs Feedforward model using STGCN temporal layer on $C_P$ prediction. The lower surface of the wing is shown. LE: Leading Edge. TE: Trailing Edge. Dash-dot line indicates the symmetry plane.}
    \label{fig:FFvsARMAX_schroeder_plot_CP}
\end{figure}

\begin{figure}[!hbt]
    \centering
    \includegraphics[trim=0.2cm 0.2cm 0.2cm 0.2cm, clip, width=1\linewidth]{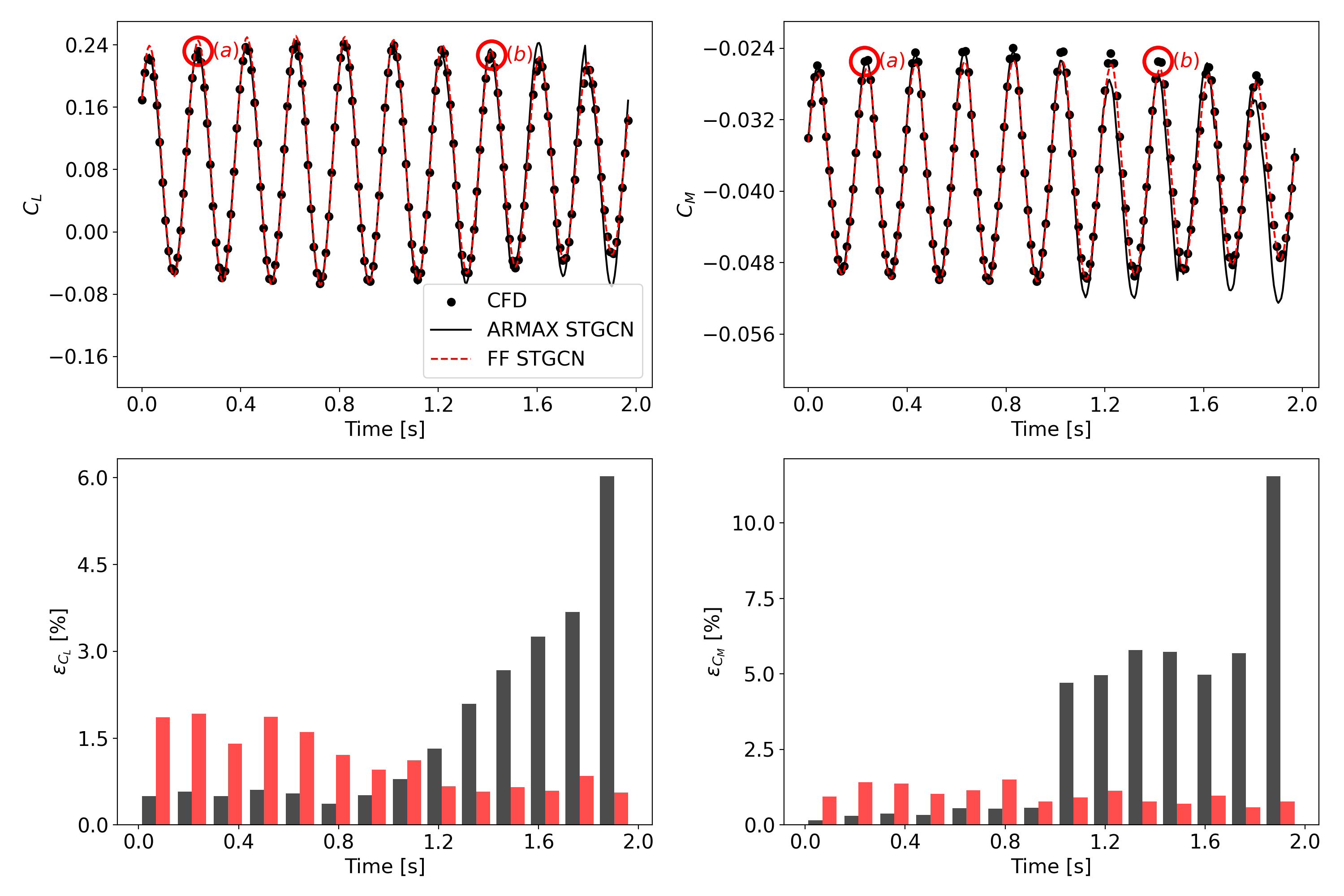}
    \caption{Validation signal 2 - SH type: ARMAX vs Feedforward model using STGCN temporal layer on $C_L$ and $C_M$ predictions. Red circles denote the points used for plotting the \( C_P \) distribution.}
    \label{fig:FFvsARMAX_harmonic_plot_CL_CM}
\end{figure}

These trends are even more evident with the SH signal, which features rapid oscillations. Figure~\ref{fig:FFvsARMAX_harmonic_plot_CL_CM} shows that the feedforward model handles these high-frequency aerodynamic variations more effectively, with significantly fewer phase and amplitude errors than the ARMAX model. The ARMAX model, due to its time-marching scheme, exhibits a greater sensitivity to reduced frequency. At higher frequencies, errors accumulate faster as small inaccuracies from previous steps compound. Once the ARMAX model switches from using ground-truth inputs to its own predictions, it fails to keep pace with the rapid changes in aerodynamic forces, resulting in larger phase lags and more significant deviations from the reference data. In contrast, as the feedforward model relies only on the wing spatial coordinates and prescribed motions at previous timesteps (without feeding back its own predictions), it results in a more stable error profile and in a more reliable and accurate predictions of unsteady phenomena.

% This error accumulation can overshadow the physical phenomena being modeled, reducing predictive accuracy.
 
\begin{figure}[!hbt]
    \centering
    \includegraphics[trim=0.2cm 0.2cm 0.2cm 0.2cm, clip, width=0.5\linewidth]{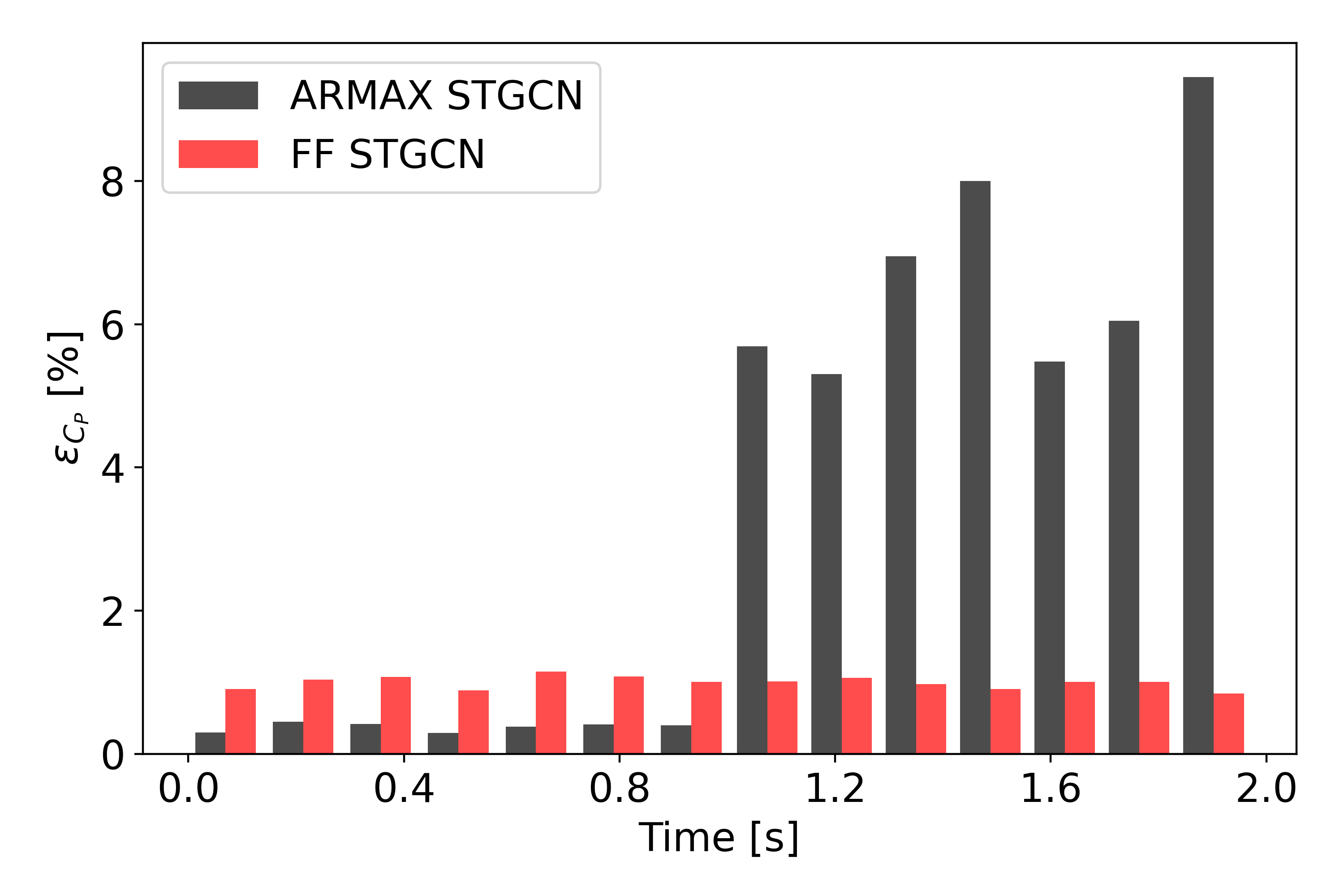}
    \caption{Validation signal 2 - SH type: Evolution of MAPE of $C_P$ with ARMAX and Feedforward model.}
    \label{fig:FFvsARMAX_harmonic_plot_MAPE_CP}
\end{figure}

As shown in Figure~\ref{fig:FFvsARMAX_harmonic_plot_MAPE_CP}, the feedforward model consistently outperforms the ARMAX model in terms of \texttt{MAPE} on \(C_P\) for the SH signal. The ARMAX model error accumulation is particularly pronounced in more dynamic conditions, such as rapid oscillations, which results in significantly higher \texttt{MAPE}. Figure~\ref{fig:FFvsARMAX_harmonic_plot_CP} further supports this observation, showing that the feedforward model is better at predicting the \(C_P\) distribution, where the ARMAX model struggles due to the accumulation of prediction errors. However, it is important to note that when the ARMAX model is feeded with ground-truth \(C_P\) values, it can produce very accurate predictions, outperforming the feedforward model. This highlights the ARMAX potential in study cases where error on pressure values does not accumulate too fast, as in systems with lower reduced frequency. Also, insufficient convergence of the CFD solution may affect error accumulation in regions with complex flow patterns, suggesting that improving solution stability may help mitigate this problem.

\begin{figure}[!hbt]
    \centering
    \includegraphics[trim=0.2cm 0.2cm 0.2cm 0.2cm, clip, width=0.8\linewidth]{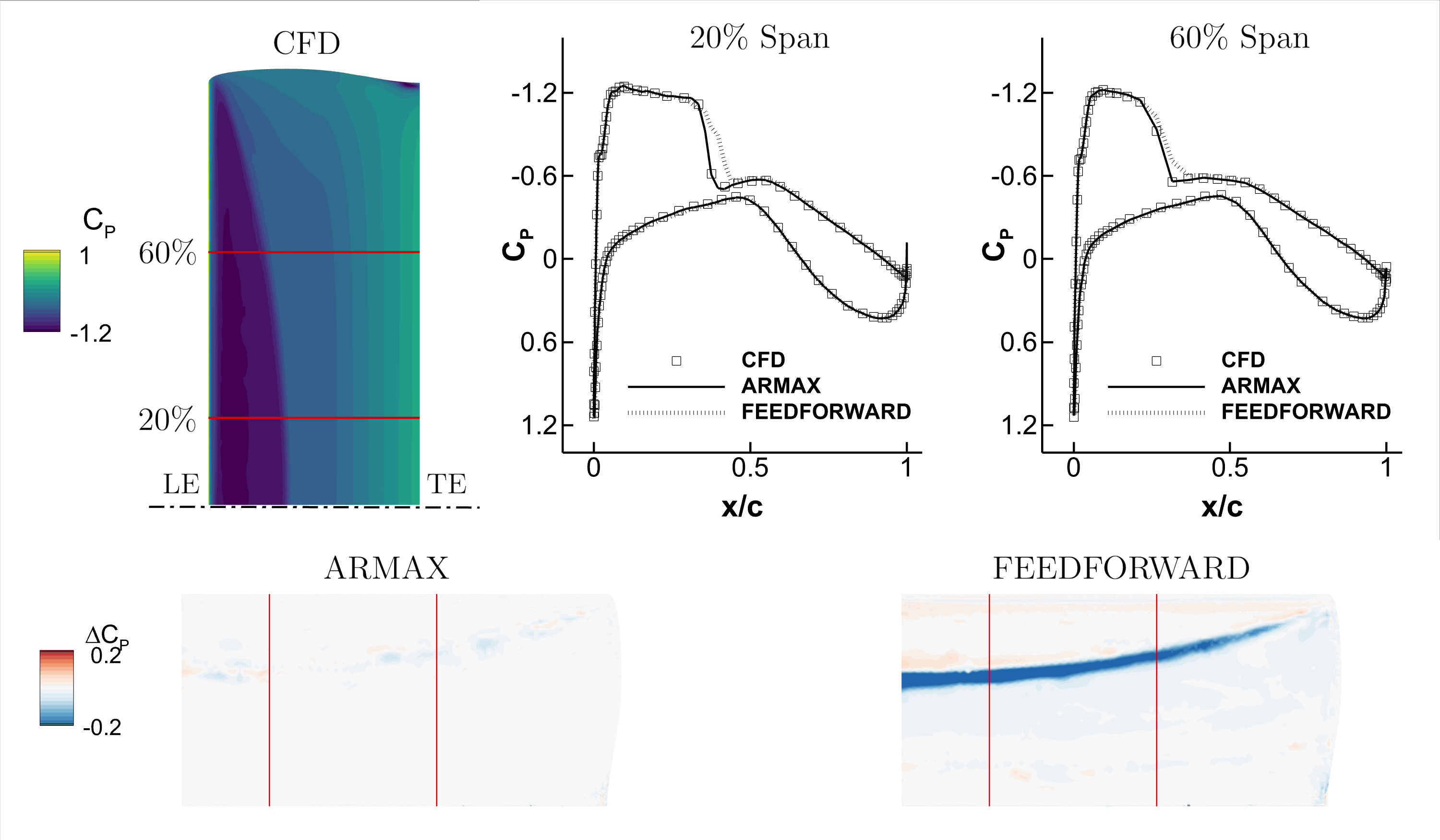}
    \caption*{Time Instance (a)}
\end{figure}
\begin{figure}[!hbt]
    \centering
    \includegraphics[trim=0.2cm 0.2cm 0.2cm 0.2cm, clip, width=0.8\linewidth]{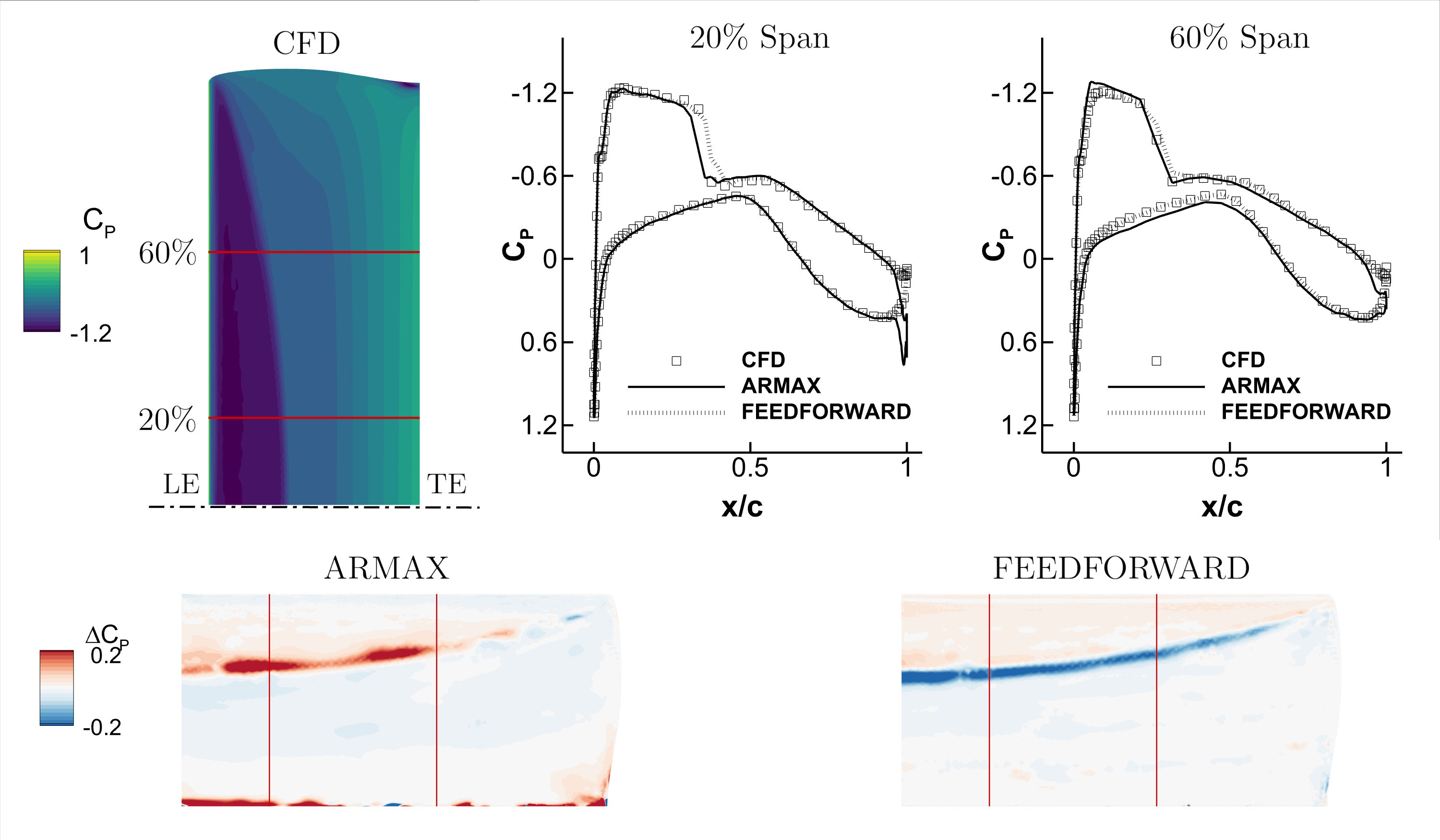}
    \caption*{Time Instance (b)}
\end{figure}
\begin{figure}[!hbt]
    \centering
    \caption{Validation signal 2 - SH type: ARMAX vs Feedforward model using STGCN temporal layer on $C_P$ prediction. The upper surface of the wing is shown. LE: Leading Edge. TE: Trailing Edge. Dash-dot line indicates the symmetry plane.}
    \label{fig:FFvsARMAX_harmonic_plot_CP}
\end{figure}

\subsection{Computing Cost Analysis}

A comprehensive analysis of computational costs was performed to compare the efficiency of the proposed model with that of a higher-order approach, as shown in Table~\ref{tab_cpu_time_comparison}. A single CFD run on a high-performance computing system, using an Intel Skylake-based architecture with 3 nodes and 40 CPU cores per node, typically requires around 6,000 CPU hours. Generating the entire dataset demands approximately 75,000 CPU hours. In contrast, the proposed framework allows predictions for a single sample on a local machine with a NVIDIA RTX A4000 GPU to be completed in roughly two minutes, leading to computational savings of over 99\%. Nonetheless, it is important to acknowledge the substantial computational effort involved in generating the high-fidelity simulations used to create the dataset. This underscores the need for strategies that reduce the amount of training data required to develop an accurate model.

% The training of the GST GraphNet model was conducted on an Intel XEON W-2255 CPU and a NVIDIA RTX A4000 GPU.

\begin{table}[!htb]
\small
\centering
\begin{tabular}{c c c c c c c}
\hline 
\hline 
\multicolumn{2}{c}{\textbf{CFD (CPU hours)}} &  \multicolumn{5}{c}{\textbf{GST GraphNet (GPU hours)}} \\
\cmidrule(lr){1-2}\cmidrule(lr){3-7}
  \multicolumn{2}{c}{Simulation}   & Pre-Trained AE & \multicolumn{2}{c}{Training} & \multicolumn{2}{c}{Prediction} \\
  (12 runs) & (1 run)  & (Optimization + Training) & (ARMAX) & (FF) & (ARMAX) & (FF) \\
\hline
75,000 &  6,000  &  42.6 &  35.2 & 33.1 & 0.03 & 0.03 \\
\hline
\hline 
\end{tabular}
\caption{Computing cost comparison between GST GraphNet model and CFD. FF: FeedForward.}
\label{tab_cpu_time_comparison}
\end{table}

\section{Conclusions}\label{sec:Conclusions}

This study introduced a framework for predicting unsteady transonic wing pressure distributions, integrating an AE architecture with GCN and graph-based temporal layers to capture time dependencies. The proposed model effectively compresses high-dimensional \( C_P \) distribution data into a lower-dimensional latent space using the AE, preserving essential features for accurate representation. The GCN layers are well-suited for handling the unstructured grids characteristic of aerodynamic data, while the temporal layers capture and leverage temporal dependencies for robust forecasting of wing \( C_P \) distributions.

Our results demonstrated that this integrated approach can achieve an accuracy comparable to traditional CFD methods, while significantly reducing computational costs. We evaluated two architectures, the feedforward model and the ARMAX model, using different temporal layers, with the STGCN layer consistently delivering the most accurate results across the validation signals. The feedforward model demonstrated clear advantages in terms of predictive accuracy and stability, particularly in avoiding the error propagation inherent in the ARMAX model. By not relying on its own predictions for subsequent inputs, the feedforward model is better suited for both steady and highly dynamic aerodynamic conditions, showing greater accuracy in predicting complex flow features, such as shock waves and flow separation. The ARMAX model, while capable of capturing general trends, is more prone to error accumulation, particularly in scenarios with high-frequency oscillations or rapid changes in aerodynamic forces. Nonetheless, when using ground-truth inputs, the ARMAX model can yield highly accurate results, underscoring its potential in scenarios with reliable data inputs.

% The study highlights the potential of combining AE, GCNs, and temporal layers to address the complexities of aerodynamic analyses. This approach offers a scalable and efficient solution for predicting unsteady aerodynamic phenomena, making it a valuable tool for future aerodynamic research and engineering applications.

Future work will focus on extending the framework to different flight regimes to validate its adaptability to a wider range of aerodynamic conditions. Additionally, scalability to larger grids and different graph structures will be explored. GCNs inherently support various graph configurations, but expanding the model to handle new spatial structures or larger meshes may require techniques like subgraph splitting or padding to ensure stable performance. As grid size increases, deeper networks and additional pooling layers will be necessary to capture long-range dependencies efficiently, while maintaining computational feasibility. 
% These improvements will enhance the model robustness and applicability across more complex and diverse aerodynamic scenarios.

\section*{Acknowledgement}

This work was supported by Digitalization Initiative of the Zurich Higher Education Institutions (DIZH) grant from Zurich University of Applied Sciences (ZHAW). The authors also acknowledge the University of Southampton for granting access to the IRIDIS High Performance Computing Facility and its associated support services.

\appendix

\section{Schroeder-Phased Harmonic Signal Formulation} \label{app:schroder_signal}

The Schroeder-phased harmonic signal is utilized in this study to improve the robustness and generalizability of the model by covering a wide frequency spectrum. These signals are constructed by summing sinusoidal components, where the phases are optimized to minimize the overall peak amplitude. This results in an evenly distributed energy spectrum, which is advantageous for training the model to handle various frequency interactions and reduces the risk of overfitting.  To cover this broad frequency range with minimal peak amplitude, a total of 9 harmonics is selected, with $M=9$.

Both damped Schroder-phased harmonic (DS) and undamped Schroder-phased harmonic (US) signals are used to model the wing displacement, whether it be pitch 
$\theta(t)$ or plunge $\xi(t)$. The US signal uniformly distributes energy across the frequency spectrum and is defined as:

\begin{equation}
\theta_{US}(t) = \sum_{m=1}^{M} a_m \sin\left((m+1) \omega_m t + \phi_m\right)
\end{equation}

where \( a_m \) represents the amplitude of the \( m \)-th component, \( \omega_m \) denotes the angular frequency, and \( \phi_m \) corresponds to the phase of the \( m \)-th sinusoidal component. %The US signal ensures a consistent energy distribution across all frequencies.

For transient response analysis, the DS signal simulates amplitude decay over time, incorporating a damping function:

\begin{equation}
\theta_{DS}(t) = \sum_{m=1}^{M} \left( \left( \frac{a_{\text{end}} - a_{0}}{t_{\text{end}} - t_{0}} (t - t_{0}) + a_{0} \right) \sin\left((m+1) \omega_m t + \phi_m\right) \right)
\end{equation}

where \( a_{0} \) and \( a_{\text{end}} \) denote the initial and final amplitudes, respectively, and \( t_{0} \) and \( t_{\text{end}} \) are the corresponding time intervals. The damping is designed such that at the final time step, the amplitude is reduced to \( 0.1 \, a_{0} \), ensuring transient behaviors are effectively captured.

The phases \( \phi_m \) are calculated to minimize constructive interference between the sinusoidal components, flattening the overall spectrum:

\begin{equation}
\phi_m = -\frac{m(m+1)\pi}{M}
\end{equation}

% The combination of damped and undamped signals allows for a comprehensive evaluation of the model ability to predict both steady-state and transient aerodynamic responses, thus enhancing the model robustness and applicability to a wide range of aerodynamic phenomena.

\section{Models Architecture} %\label{app:optarch}

This section outlines the architecture and training process for both the ARMAX and feedforward models used in this study, as detailed in Tables~\ref{tab:armax_architecture} and \ref{tab:feedforward_architecture}. The ARMAX model in Table~\ref{tab:armax_architecture} combines autoregressive components with GCN layers and STGCN temporal layer to capture both spatial and temporal dynamics, featuring 5,775,023 trainable weights. In contrast, the feedforward model in Table~\ref{tab:feedforward_architecture} avoids using previous predictions, which helps prevent error accumulation over time. This model has 1,962,111 trainable weights. Both models utilize a pre-trained AE for dimensionality reduction. Specifically, the Encoding A and Decoding layers in both architectures are optimized based on the pre-trained AE, while Encoding B mirrors the structure of Encoding A, ensuring consistent feature extraction across different model variants. Additionally, the concatenation block in both models concatenates the encodings from the previous three timesteps, enabling the temporal layer to effectively capture and process the sequential dependencies within the data.

During the backpropagation phase, the ADAptive Moment Estimation (Adam) optimizer \cite{kingma2014adam} was employed to fine-tune the neural network weights and minimize the \texttt{MAE} loss function. The learning rate was set to 0.001. A batch size $m$ of 1 was found to yield the most accurate results. The training process was carried out over 50 epochs.

\begin{table}[!htb]
\small
\begin{tabular}{ccccccccc}
\cline{1-5} \cline{7-9}
\multicolumn{1}{|l|}{\multirow{9}{*}{\rotatebox[origin=c]{90}{\textbf{Encoding A}}}} & \multicolumn{3}{c}{\textbf{Layer Type}} & \textbf{Output Size} & \multicolumn{1}{c|}{\multirow{9}{*}{}} & \multicolumn{1}{c|}{\multirow{9}{*}{\rotatebox[origin=c]{90}{\textbf{Encoding B}}}} & \textbf{Layer Type} & \multicolumn{1}{c}{\textbf{Output Size}} \\ \cline{2-5} \cline{8-9} 
\multicolumn{1}{|l|}{}                                     & \multicolumn{3}{c}{Input}               & $m\times 3\times 86840\times 8$         & \multicolumn{1}{c|}{}                  & \multicolumn{1}{c|}{}                                     & Input               & $m\times 3\times 86840\times 1$                             \\ \cline{2-5} \cline{8-9} 
\multicolumn{1}{|l|}{}                                     & \multicolumn{3}{c}{GCN}                 & $m\times 3\times 86840\times 256$         & \multicolumn{1}{c|}{}                  & \multicolumn{1}{c|}{}                                     & GCN                 & $m\times 3\times 86840\times 256$                             \\
\multicolumn{1}{|l|}{}                                     & \multicolumn{3}{c}{GCN}                 & $m\times 3\times 86840\times 224$         & \multicolumn{1}{c|}{}                  & \multicolumn{1}{c|}{}                                     & GCN                 & $m\times 3\times 86840\times 224$                             \\
\multicolumn{1}{|l|}{}                                     & \multicolumn{3}{c}{GCN}                 & $m\times 3\times 86840\times 96$         & \multicolumn{1}{c|}{}                  & \multicolumn{1}{c|}{}                                     & GCN                 & $m\times 3\times 86840\times 96$                             \\ \cline{2-5} \cline{8-9} 
\multicolumn{1}{|l|}{}                                     & \multicolumn{3}{c}{Pooling 1}           & $m\times 3\times 28600\times 96$         & \multicolumn{1}{c|}{}                  & \multicolumn{1}{c|}{}                                     & Pooling 1           & $m\times 3\times 28600\times 96$                            \\ \cline{2-5} \cline{8-9} 
\multicolumn{1}{|l|}{}                                     & \multicolumn{3}{c}{GCN}                 & $m\times 3\times 28600\times 64$         & \multicolumn{1}{c|}{}                  & \multicolumn{1}{c|}{}                                     & GCN                 & $m\times 3\times 28600\times 64$                             \\ \cline{2-5} \cline{8-9} 
\multicolumn{1}{|l|}{}                                     & \multicolumn{3}{c}{Pooling 2}           & $m\times 3\times 9600\times 64$         & \multicolumn{1}{c|}{}                  & \multicolumn{1}{c|}{}                                     & Pooling 2           & $m\times 3\times 9600\times 64$                             \\ \cline{2-5} \cline{8-9} 
\multicolumn{1}{|l|}{}                                     & \multicolumn{3}{c}{GCN}                 & $m\times 3\times 9600\times 368$         & \multicolumn{1}{c|}{}                  & \multicolumn{1}{c|}{}                                     & GCN                 & $m\times 3\times 9600\times 368$                            \\ \hline \hline
\multicolumn{9}{c}{\textbf{Concatenate Block} -- Output: $m\times 3 \times 9600\times 736$}                                                                                                                                                                                                                                                   \\ \hline \hline
\multicolumn{9}{c}{\textbf{Temporal Layer} -- Output: $m\times 9600\times 368$}                                                                                                                                                                                                                                                       \\ \hline \hline
\multicolumn{1}{|l|}{\multirow{9}{*}{\rotatebox[origin=c]{90}{\textbf{Decoding}}}}   & \multicolumn{6}{c}{\textbf{Layer Type}}                                                                                                                             & \multicolumn{2}{c}{\textbf{Output Size}}                            \\ \cline{2-9} 
\multicolumn{1}{|l|}{}                                     & \multicolumn{6}{c}{GCN}                                                                                                                                             & \multicolumn{2}{c}{$m\times 9600\times 368$}                                \\ \cline{2-9} 
\multicolumn{1}{|l|}{}                                     & \multicolumn{6}{c}{Unpooling 2}                                                                                                                                     & \multicolumn{2}{c}{$m\times 28600\times 368$}                                \\ \cline{2-9} 
\multicolumn{1}{|l|}{}                                     & \multicolumn{6}{c}{GCN}                                                                                                                                             & \multicolumn{2}{c}{$m\times 28600\times 64$}                                \\ \cline{2-9} 
\multicolumn{1}{|l|}{}                                     & \multicolumn{6}{c}{Unpooling 1}                                                                                                                                     & \multicolumn{2}{c}{$m\times 86840\times 64$}                                \\ \cline{2-9} 
\multicolumn{1}{|l|}{}                                     & \multicolumn{6}{c}{GCN}                                                                                                                                             & \multicolumn{2}{c}{$m\times 86840\times 96$}                                \\
\multicolumn{1}{|l|}{}                                     & \multicolumn{6}{c}{GCN}                                                                                                                                             & \multicolumn{2}{c}{$m\times 86840\times 224$}                                \\
\multicolumn{1}{|l|}{}                                     & \multicolumn{6}{c}{GCN}                                                                                                                                             & \multicolumn{2}{c}{$m\times 86840\times 256$}                                \\ \cline{2-9} 
\multicolumn{1}{|l|}{}                                     & \multicolumn{6}{c}{Output}                                                                                                                                          & \multicolumn{2}{c}{$m\times 86840\times 1$}                               \\ \hline
\end{tabular}
\caption{Layer structure and output dimensions for the ARMAX model, detailing the two encodings, temporal, and decoding layers used for predicting pressure distribution.}
\label{tab:armax_architecture}
\end{table}

\begin{table}[!htb]
    \centering
    \small
    \begin{tabular}{ccccc}
        \cline{1-5}
        \multicolumn{1}{|l|}{\multirow{9}{*}{\rotatebox[origin=c]{90}{\textbf{Encoding A}}}} & \multicolumn{3}{c}{\textbf{Layer Type}} & \textbf{Output Size} \\ \cline{2-5}
        \multicolumn{1}{|l|}{} & \multicolumn{3}{c}{Input} & $m\times 3\times 86840\times 8$ \\ \cline{2-5}
        \multicolumn{1}{|l|}{} & \multicolumn{3}{c}{GCN} & $m\times 3\times 86840\times 256$ \\
        \multicolumn{1}{|l|}{} & \multicolumn{3}{c}{GCN} & $m\times 3\times 86840\times 224$ \\
        \multicolumn{1}{|l|}{} & \multicolumn{3}{c}{GCN} & $m\times 3\times 86840\times 96$ \\ \cline{2-5}
        \multicolumn{1}{|l|}{} & \multicolumn{3}{c}{Pooling 1} & $m\times 3\times 28600\times 96$ \\ \cline{2-5}
        \multicolumn{1}{|l|}{} & \multicolumn{3}{c}{GCN} & $m\times 3\times 28600\times 64$ \\ \cline{2-5}
        \multicolumn{1}{|l|}{} & \multicolumn{3}{c}{Pooling 2} & $m\times 3\times 9600\times 64$ \\ \cline{2-5}
        \multicolumn{1}{|l|}{} & \multicolumn{3}{c}{GCN} & $m\times 3\times 9600\times 368$ \\ \hline\hline
        \multicolumn{5}{c}{\textbf{Temporal Layer} -- Output: $m\times 9600\times 368$} \\ \hline\hline
        \multicolumn{1}{|l|}{\multirow{9}{*}{\rotatebox[origin=c]{90}{\textbf{Decoding}}}} & \multicolumn{3}{c}{\textbf{Layer Type}} & \textbf{Output Size} \\ \cline{2-5}
        \multicolumn{1}{|l|}{} & \multicolumn{3}{c}{GCN} & $m\times 9600\times 368$ \\ \cline{2-5}
        \multicolumn{1}{|l|}{} & \multicolumn{3}{c}{Unpooling 2} & $m\times 28600\times 368$ \\ \cline{2-5}
        \multicolumn{1}{|l|}{} & \multicolumn{3}{c}{GCN} & $m\times 28600\times 64$ \\ \cline{2-5}
        \multicolumn{1}{|l|}{} & \multicolumn{3}{c}{Unpooling 1} & $m\times 86840\times 64$ \\ \cline{2-5}
        \multicolumn{1}{|l|}{} & \multicolumn{3}{c}{GCN} & $m\times 86840\times 96$ \\
        \multicolumn{1}{|l|}{} & \multicolumn{3}{c}{GCN} & $m\times 86840\times 224$ \\
        \multicolumn{1}{|l|}{} & \multicolumn{3}{c}{GCN} & $m\times 86840\times 256$ \\ \cline{2-5}
        \multicolumn{1}{|l|}{} & \multicolumn{3}{c}{Output} & $m\times 86840\times 1$ \\ \hline
    \end{tabular}
\caption{Layer structure and output dimensions for the Feedforward model, detailing the encoding, temporal, and decoding layers used for predicting pressure distribution.}
\label{tab:feedforward_architecture}
\end{table}

\bibliographystyle{elsarticle-num-names} 
\bibliography{cas-refs}

\end{document}